\documentclass{article}
\PassOptionsToPackage{table}{xcolor}
\usepackage{iclr2027_conference,times}
\usepackage[T1]{fontenc}
\usepackage{iftex}
\ifPDFTeX
  \usepackage[utf8]{inputenc}
\else\ifXeTeX
  \AtBeginDvi{}
\fi\fi
\usepackage{latexsym,amsmath}
\usepackage{microtype,inconsolata}
\usepackage{graphicx,enumitem,booktabs,multirow,makecell,array}
\usepackage[table]{xcolor}
\usepackage{algorithm,algpseudocode}
\usepackage{pifont}
\usepackage{url,hyperref}
\usepackage{cleveref}
\usepackage{placeins}
\usepackage{wrapfig}
\usepackage{caption}
\crefname{table}{Table}{Tables}
\crefname{figure}{Figure}{Figs.}
\crefname{section}{Sec.}{Secs.}
\crefname{equation}{Eq.}{Eqs.}
\crefname{algorithm}{Alg.}{Algs.}
\crefname{appendix}{App.}{Apps.}
\newcommand{\cmark}{\ding{51}}
\newcommand{\xmark}{\ding{55}}

\title{VideoWeaver: Evaluating and Evolving \\ Skills for Agentic Long Video Generation}

\author{%
  \normalfont
    Jianhui Wei\textsuperscript{1,2,*,\S} \quad
    Yan Zhang\textsuperscript{2,*,\ddag} \quad
    Jie Tan\textsuperscript{1,*} \quad
    Hengchuan Zhu\textsuperscript{1,*} \quad
    Xiaotian Zhang\textsuperscript{1} \\[0.4em]
    Ziyi Chen\textsuperscript{2} \quad
    Daoan Zhang\textsuperscript{2} \quad
    Wei Xu\textsuperscript{2,\ddag} \quad
    Yeying Jin\textsuperscript{3} \quad
    Zuozhu Liu\textsuperscript{1,\dag} \\[0.6em]
  \textsuperscript{1}Zhejiang University \quad
  \textsuperscript{2}ByteDance \\[0.2em]
  \textsuperscript{3}National University of Singapore \\[0.3em]
  \texttt{\{jianhui1.24,zuozhuliu\}@intl.zju.edu.cn}
}

\iclrfinalcopy
\hypersetup{
  pdftitle={VideoWeaver: Evaluating and Evolving Skills for Agentic Long Video Generation},
  pdfauthor={Jianhui Wei, Yan Zhang, Jie Tan, Hengchuan Zhu, Xiaotian Zhang, Ziyi Chen, Daoan Zhang, Wei Xu, Yeying Jin, Zuozhu Liu}
}

\makeatletter
\newcommand{\finishresultwrap}{%
  \par
  \ifnum\c@WF@wrappedlines>2
    \vspace{\dimexpr\c@WF@wrappedlines\baselineskip-2\baselineskip\relax}%
  \fi
  \WFclear
}
\makeatother

\begin{document}
\maketitle
\lhead{Preprint}
\raggedbottom

\ificlrfinal
\renewcommand{\thefootnote}{\fnsymbol{footnote}}
\footnotetext[1]{\ Equal contributions.}
\footnotetext[2]{\ Corresponding author.}
\footnotetext[3]{\ Project Leader.}
\footnotetext[4]{\ Work done during internship at ByteDance.}
\renewcommand{\thefootnote}{\arabic{footnote}}
\setcounter{footnote}{0}
\fi

\begin{abstract}
Agentic long video generation requires planning, tool orchestration, and
cross-clip coordination over a long horizon. Most existing video agents either
rely on static, human-crafted workflows, which require substantial manual
effort and poorly adapt across tasks, or iteratively refine the output of the
current task without persistently distilling execution experience into
reusable skills for future tasks.We introduce VideoWeaver,
an agent harness and benchmark that evaluates
and evolves skills for long video generation. Given a single high-level
instruction, an agent dynamically composes foundation skills into its own
workflow rather than following a predefined pipeline. We construct a benchmark of 16 task categories
and 285 cases, with references spanning text, image, audio, video, and their
combinations. We further propose an evidence-grounded agent-as-judge that
inspects both the execution trace and the final video to diagnose process and
output failures. Based on this feedback, our evolution algorithm
progressively refines category-level composition and creator skills, allowing
recurring experience to guide dynamically constructed workflows for unseen
cases. Experiments show that explicit composition skills improve the generation
process over foundation skills alone, while skill evolution further improves
output quality and generalizes to unseen cases. Incorporating judge feedback
yields additional gains, especially on output metrics, and the agent-as-judge
aligns well with human, particularly on process metrics.
Code is available at \url{https://github.com/JianhuiWei7/VideoWeaver}.
\end{abstract}

\section{Introduction}

Long video generation is advancing rapidly along two complementary lines. End-to-end
models~\citep{cui2026lollongerlongerscaling,streamingt2v,helios}
extend a single generative process, while agentic
systems~\citep{huang2026vimaxagenticvideogeneration,univa,codirector} orchestrate generation tools across many
steps. The agentic line suits the complex, multi-shot tasks we target, which require
planning over a long horizon, coordinating multimodal references such as images, audio,
and video, invoking different generation and editing tools, and maintaining
narrative, visual, and audio consistency across many clips. Such agentic systems already address diverse
scenarios, from cinematic storytelling~\citep{script,Hollywood} and personalized
vlogs~\citep{Personavlog} to music videos~\citep{Automv} and animation~\citep{Animaker}.
Representative frameworks such as
ViMax~\citep{huang2026vimaxagenticvideogeneration}, StoryAgent~\citep{storyagent}, UniVA~\citep{univa}, and
Co-Director~\citep{codirector} demonstrate this promise. However, they rely on
purpose-built orchestration mechanisms that require substantial human design
and do not persistently improve from experience across cases. Meanwhile,
general-purpose agent frameworks such as Claude Code~\citep{anthropic-claude-code-2026},
Codex~\citep{openai-codex-2026}, and OpenClaw~\citep{openclaw-2026} pair a reasoning
backbone with a tool-augmented runtime and show strong tool-use and orchestration. Nevertheless, their execution experience is not automatically distilled into
reusable skills for future tasks. Whether such general-purpose harnesses can
handle long video generation while improving their workflows from feedback has not been systematically explored.

To study this, we introduce VideoWeaver, a framework for long video generation built on a general-purpose agent harness. In our setting, a backbone
model~\citep{openai-gpt55-2026,bytedance-seed2-2026,deepseek-v4-2026} is equipped with a set of
\emph{foundation skills}: self-contained, independently invocable capabilities for video,
image, and audio generation, understanding, and media processing. Given a single
high-level instruction, the agent composes these foundation skills into its own
\emph{composition skill}, a high-level procedural workflow, and executes it through
multi-step skill invocation and clip composition rather than following a predefined
pipeline. To make this setting concrete and measurable, we construct a benchmark of 16
task categories and 285 cases, with references spanning text, image, audio, video, and
their combinations (\cref{fig:dataset_distribution}). We use in-distribution train and test splits and hold out categories for out-of-distribution evaluation.

\begin{table}[t]
\caption{
Evaluation coverage and scale across video-generation, production, and general skill benchmarks; detailed statistics and counting conventions are provided in \cref{app:benchmark_statistics}.
}
\label{tab:metric_coverage}
\label{tab:benchmark_scale}
\centering
\scriptsize
\setlength{\tabcolsep}{2.5pt}
\renewcommand{\arraystretch}{1.00}
\begin{tabular}{@{}>{\raggedright\arraybackslash}p{0.36\textwidth}>{\raggedright\arraybackslash}p{0.30\textwidth}>{\raggedright\arraybackslash}p{0.13\textwidth}cc@{}}
\toprule
\textbf{Benchmark / Method} & \textbf{Domain} & \textbf{Scale} & \textbf{Process} & \textbf{Output} \\
\midrule
\multicolumn{5}{@{}l}{\textit{Video-generation and production benchmarks}} \\
StoryAgent~\citep{storyagent} & Customized storytelling & 90 samples & \xmark & \cmark~(9) \\
GenAD-Bench~\citep{codirector} & Video advertising & 400 scenarios & \xmark & \cmark~(4) \\
ThemeVlogEval~\citep{Personavlog} & Personalized Vlog generation & 10 portraits & \xmark & \cmark~(12) \\
AutoMV~\citep{Automv} & Music-video generation & 30 songs & \xmark & \cmark~(13) \\
ST-Bench~\citep{zhang2025storymemmultishotlongvideo} & Multi-shot storytelling & 300 prompts & \xmark & \cmark~(5) \\
StreamingT2V~\citep{streamingt2v} & Long text-to-video generation & 50 prompts & \xmark & \cmark~(3) \\
HeliosBench~\citep{helios} & Real-time long-video generation & 240 prompts & \xmark & \cmark~(9) \\
AniEval~\citep{Animaker} & Multi-shot animation & 10 stories & \xmark & \cmark~(14) \\
UniVA-Bench~\citep{univa} & Agentic video intelligence & 130 cases & \cmark~(3) & \cmark~(6) \\
AgenticVBench~\citep{cao2026agenticvbenchaiagentscomplete} & Video post-production & 100 tasks & \xmark & \cmark~(Var.) \\
\midrule
\multicolumn{5}{@{}l}{\textit{General skill benchmarks and methods}} \\
SkillsBench~\citep{li2026skillsbenchbenchmarkingagentskills} & General & 87 tasks & \xmark & \cmark~(1) \\
\multirow{3}{*}{\makecell[l]{SkillOpt\\\citep{yang2026skilloptexecutivestrategyselfevolving}}} & ALFWorld & 191 tasks & \xmark & \cmark~(1) \\
& LiveMathBench & 177 tasks & \xmark & \cmark~(1) \\
& OfficeQA & 246 tasks & \xmark & \cmark~(1) \\
SkillFlow~\citep{skillflow-2026} & General & 166 tasks & \cmark~(5) & \cmark~(1) \\
ContinualSkill\allowbreak Bench~\citep{continualskillbench-2026} & General & 500 tasks & \xmark & \cmark~(5) \\
\midrule
\rowcolor{gray!12}
\textbf{VideoWeaver (Ours)} & \textbf{Agentic long-video generation} & \textbf{285 cases} & \textbf{\cmark~(6)} & \textbf{\cmark~(7)} \\
\bottomrule
\end{tabular}
\vspace{-10pt}
\end{table}
Evaluating such agents is itself difficult, because errors can occur at any stage of the
generation process and affect the quality of the final video. A plan may be flawed, a tool call may
fail, or an intermediate artifact may be corrupted in ways the final clip does not
reveal. Most existing video benchmarks~\citep{vbench-2023,videophy-2024,visionreward-2024,videobench-2025,univbench-2026,clvgbench-2026}
focus on the final output (\cref{tab:metric_coverage}) and provide
limited coverage of process-level failures. We therefore propose a skill-augmented \emph{agent-as-judge}
that inspects both the execution trace and the final video (\cref{fig:evaluation_harness}).
It invokes tools to extract hard evidence such as video metadata and intermediate files,
then returns a score and textual feedback for each of the six process and seven output metrics, with
emphasis on cross-clip consistency. Using this feedback as a refinement signal, we further
design a skill evolution algorithm (\cref{fig:evolution_pipeline}) that progressively
refines category-level composition and creator skills and merges the resulting creator
skills into a single one.

We conduct experiments across multiple agent frameworks and foundation models. We find that performance varies notably across harness and model choices, explicit composition skills clearly improve the generation process over foundation skills alone, and skill evolution further improves output quality while generalizing to unseen cases. Incorporating judge feedback during evolution yields additional gains, especially on output metrics. In addition, a human study confirms that our agent-as-judge aligns well with human judgments. Overall, we make the following contributions.

\begin{itemize}[leftmargin=*, itemsep=1pt, topsep=2pt, parsep=0pt]
    \item We introduce VideoWeaver, a framework for studying agentic long video generation, together with a benchmark spanning 16 task categories and 285 cases.

    \item We propose an evidence-grounded agent-as-judge that diagnoses
    execution traces and final outputs, providing feedback to guide skill evolution.

    \item We develop a evolution algorithm that distills evaluation
    feedback into reusable composition and creator skills, improving output
    quality and generalizing to unseen cases.
\end{itemize}

\section{Related Work}

\subsection{Long Video Generation}

Video generation models such as Sora~\citep{videoworldsimulators2024} and Veo~\citep{google_veo_2024}
produce high-fidelity short clips but still struggle with coherence and temporal drift over long
durations. Recent long-video methods address this by improving autoregressive generation and
memory handling: StreamingT2V~\citep{streamingt2v} uses short-term conditioning and long-term
appearance preservation for smooth chunk transitions; LoViC~\citep{jiang2025lovicefficientlongvideo}
compresses video-text context for efficient segment-wise generation; Stable Video
Infinity~\citep{li2025stablevideoinfinityinfinitelength} recycles autoregressive errors during
training to reduce drift; Helios~\citep{helios} combines drift-aware
training with context compression for real-time minute-scale generation; and
LoL~\citep{cui2026lollongerlongerscaling} mitigates sink-collapse with training-free RoPE jitter
to enable hour-scale streaming generation.

A second line reframes long video generation as agentic orchestration, where a language model
plans, decomposes, and refines across production stages, with applications to cinematic
storytelling~\citep{script,Hollywood}, personalized vlogs~\citep{Personavlog}, music
videos~\mbox{\citep{Automv}}, and animation~\citep{Animaker}. ViMax~\citep{huang2026vimaxagenticvideogeneration}
uses hierarchical narrative planning for
long-form generation, while UniVA~\citep{univa} decomposes user requests into structured
video-processing steps. 
StoryAgent~\citep{storyagent} and Co-Director~\citep{codirector} add critique, persona
specialization, and feedback-based revision. FRAMEWORKERS~\citep{li2026frameworkersdynamicmultiagentframework}
further supports dynamic sub-agent scheduling. These purpose-built systems motivate studying
general-purpose agents that can build and evolve their own workflows.

\subsection{Video Evaluation}

Video generation is typically evaluated by reward-model-based or LLM-as-judge methods.
Reward-based benchmarks such as VBench~\citep{vbench-2023}, VideoPhy~\citep{videophy-2024}, and
VisionReward~\citep{visionreward-2024} decompose quality into dimensions scored by specialized
models, but need many task-specific evaluators, give limited textual feedback, and can be
vulnerable to reward hacking~\citep{skalse2025definingcharacterizingrewardhacking}. LLM-as-judge benchmarks such as Video-Bench~\citep{videobench-2025}, UniVBench~\citep{univbench-2026}, and
CLVG-Bench~\citep{clvgbench-2026} return scores and feedback across richer dimensions, while other methods dynamically formulate questions and provide interpretable feedback
for multi-round visual-generation evaluation or closed-loop video prompt refinement~\citep{zhang2025evaluationagentefficientpromptable,song2026vqqaagenticapproachvideo}.
However, these methods still primarily judge the final output, offering limited diagnosis of where
a long generation workflow fails.
We instead use a skill-augmented agent-as-judge that inspects both outputs and execution traces
and invokes tools to extract hard evidence such as video metadata and intermediate files.

\subsection{Self-Evolving Agent Skills}

A growing body of work studies whether LLM agents can acquire and evolve reusable skills rather
than rely on static prompts. Voyager~\citep{voyager-2023} first showed that an agent can
accumulate reusable code skills through interaction and self-reflection. Later work refines
skills from execution feedback~\citep{evoskill-2026,coevo-skills-2026,trace2skill-2026}, treats
markdown or behavioral skills as persistent, self-improving memory~\citep{memento-skills-2026,metaclaw-2026,xskill-2026,autoskill-2026,memskill-2026,memrl-2026},
and validates self-evolving skills in industrial and medical
settings~\citep{skillforge-2026,medical-imaging-agents-2026,abstral-2026}. We extend this line to agentic long video generation, where
feedback from execution traces and final videos is used to refine
reusable composition and creator skills across cases. 

\section{Task Definition}
\label{sec:dataset_eval_setup}

\subsection{Dataset Construction}
\label{sec:task_definition}



In our setting, \textbf{short-video generation} refers to tasks that can be completed with a single video-generation API call, with the user prompt and reference media supplied directly to the model. VideoWeaver instead focuses on \textbf{long-video generation}, which requires concatenating the outputs of multiple video-generation calls. Starting from a one-time user instruction, the agent plans and executes multi-step skill invocations, manages intermediate outputs, and progressively composes them into a coherent final video.

\begin{wrapfigure}[20]{R}{0.52\textwidth}
    \centering
    \includegraphics[width=\linewidth]{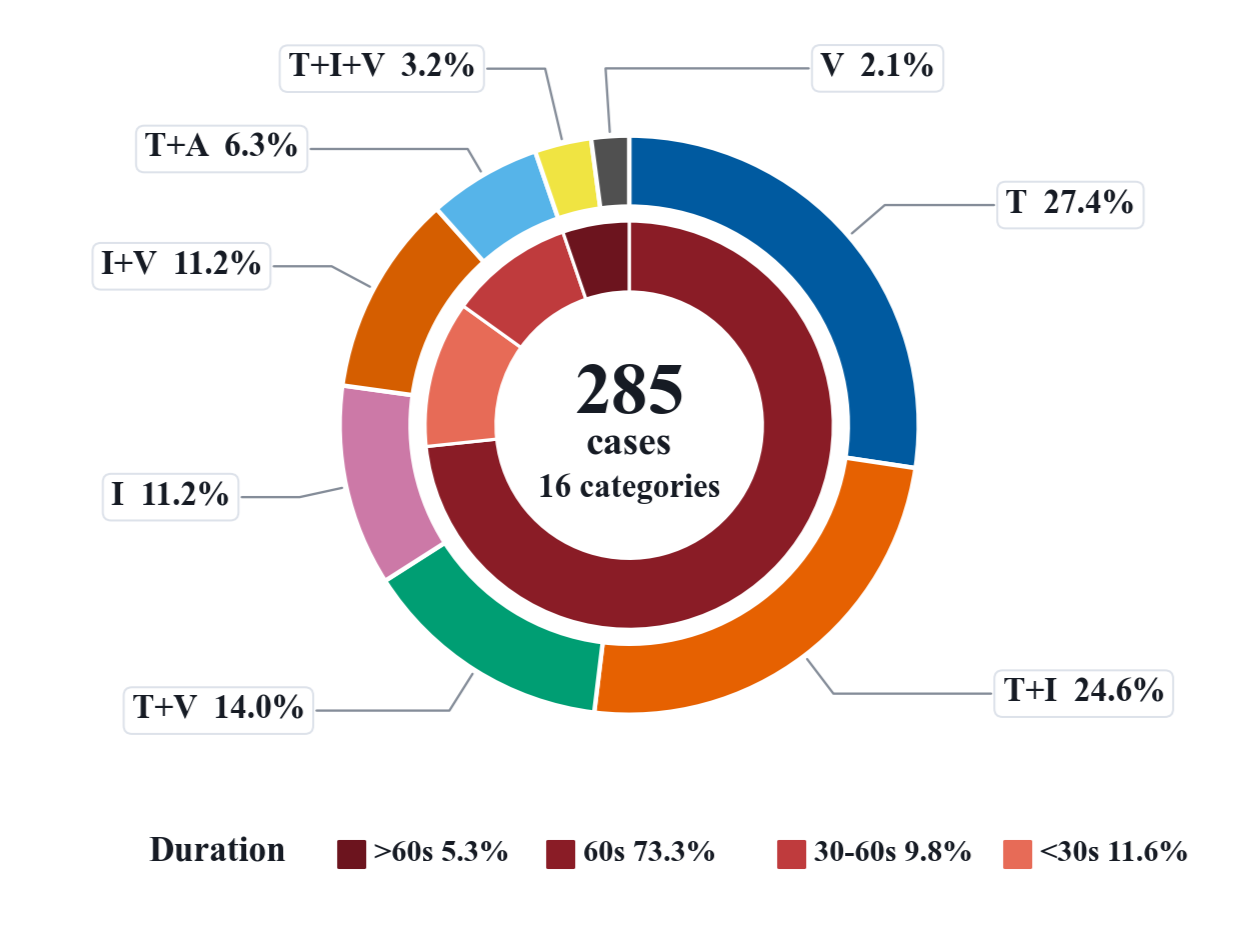}
    \caption{
     Dataset distribution of input modalities and required video durations. Modality abbreviations: T: Text, I: Image, V: Video, A: Audio.
    }
    \label{fig:dataset_distribution}
\end{wrapfigure}

We construct a dataset comprising \textbf{16 categories} of long video generation tasks and \textbf{285 cases} in total, which is of the same order of magnitude as existing benchmarks shown in
\cref{tab:benchmark_scale}. Each case includes basic textual requirements such as duration and resolution, multimodal reference materials including images, audio, videos, and their combinations, instructions on how to use these materials, and a coarse story guideline.~\cref{fig:dataset_distribution} summarizes the dataset distribution, including the modality composition of task inputs, and the distribution of the required video output duration. We further divide the dataset into training, test, and out-of-distribution (OOD) splits. Specifically, three task categories are reserved exclusively for OOD evaluation, while instances from the remaining categories are evenly partitioned into training and test sets. This setup enables the assessment of both in-distribution performance and generalization to unseen task categories. Details of task collection and filtering are provided in \cref{app:task_collection}. Representative task cases are shown in \cref{app:dataset_cases}.

\subsection{Skills Workspace}
\label{sec:foundation_composition_skills}

In our harness environment, the backbone language model (e.g., GPT-5.5~\citep{openai-gpt55-2026}, Seed2.0~\citep{bytedance-seed2-2026}) has only text generation and reasoning capabilities, but does not natively support video generation or multimodal media processing. To extend the model's ability to operate in long video generation settings, we provide a set of well-formalized foundation skills: self-contained skills that expose basic, independently invocable capabilities for video, image, and audio generation, understanding, and media processing, all of which are publicly available. Examples include generating video clips and synthesizing images. The full set of foundation skills is provided in~\cref{app:foundation_skills}. 

However, coherent long video generation requires capabilities beyond the isolated execution of foundation skills. The agent must perform long-horizon planning: decomposing a user request into intermediate objectives, selecting and sequencing appropriate foundation skills, managing intermediate artifacts across steps, and preserving narrative, visual, and audio consistency. We define \textbf{composition skills} as high-level procedural policies that specify how foundation skills are orchestrated to accomplish such long-horizon generation tasks. We further define \textbf{skill-creator} as a skill that constructs composition skills from the available foundation skills and task cases. Together, composition and creator skills provide a persistent and editable workflow representation through which execution experience can be distilled and reused across cases.

\subsection{Expert Validation}
\label{sec:expert_validation}

To validate the dataset and foundation skills, we manually construct expert composition skills for each task category through iterative generation, inspection, and refinement. This process verifies task feasibility, assesses the reliability and coverage of the foundation skills in realistic long video generation workflows, and provides a strong human-orchestrated reference baseline.

\section{Method}

\subsection{Metrics}
\label{sec:metrics}
Following prior work~\citep{lightman2023letsverifystepstep,deshpande2025trailtracereasoningagentic,barke2026agentrxdiagnosingaiagent}, we evaluate both the execution trace and the final generated video. Process metrics check execution checkpoints in the trace, such as whether the agent completes planning, input processing, and clip merging steps. Output metrics evaluate the
quality and semantic consistency of the generated artifacts, including
intermediate clips, images, and the final composed video. Unlike prior evaluations \citep{vbench-2023,videophy-2024} that mainly focus on a single video clip, we emphasize cross-clip consistency, assessing whether visual, audio, and narrative content remain consistent and coherent \textbf{across multiple clips}. Unless otherwise specified, each metric is scored on a binary scale of $\{0,1\}$ with textual feedback. Details are shown in~\cref{tab:metric_definitions}.

\begin{table}[t]
\caption{
Definitions and abbreviations of the process and output metrics used in our evaluation.
}
\label{tab:metric_definitions}
\centering
\small
\setlength{\tabcolsep}{3pt}
\renewcommand{\arraystretch}{1.05}
\begin{tabular}{@{}p{0.26\linewidth} p{0.07\linewidth} p{\dimexpr0.67\linewidth-12pt\relax}@{}}
\toprule

\textbf{Metric} & \textbf{Abb.} & \textbf{Definition} \\
\specialrule{\lightrulewidth}{0pt}{0pt}
\rowcolor{gray!12}
\multicolumn{3}{l}{\textbf{\textit{Process Metrics}}} \\
{\footnotesize Execution Error-Free Rate} & EE & Successful tool calls / total tool calls. \\
{\footnotesize Clip Merging} & CM & Whether generated clips are merged into a final long video. \\
{\footnotesize Input Processing} & IP & Whether multimodal inputs are understood and preprocessed. \\
{\footnotesize Planning} & PL & Whether the agent produces a plan before generation. \\
{\footnotesize Skill Following} & SF & Proportion of prescribed composition-skill steps correctly followed. \\
{\footnotesize Reference Asset Usage} & RAU & Whether user-provided assets are used as required. \\
\specialrule{\lightrulewidth}{0pt}{0pt}
\rowcolor{gray!12}
\multicolumn{3}{l}{\textbf{\textit{Output Metrics}}} \\
{\footnotesize Format Requirements} & FR & Whether duration, aspect ratio, and resolution are satisfied. \\
{\footnotesize Prompt Requirements} & PR & Whether the generated video satisfies instruction requirements. \\
{\footnotesize Visual Consistency} & VC & Whether characters and scenes remain consistent across clips. \\
{\footnotesize Audio Consistency} & AC & Whether speaker identity and BGM remain consistent across clips. \\
{\footnotesize Audio-Video Consistency} & AVC & Whether audio and video are synchronized. \\
{\footnotesize Plot Logic and Coherence} & PLC & Whether the story is understandable and logically connected. \\
{\footnotesize Reference Fidelity} & RF & Whether the final video reflects multimodal references. \\
\bottomrule
\end{tabular}
\end{table}

\begin{figure}[htbp]
    \centering
    \includegraphics[width=\textwidth]{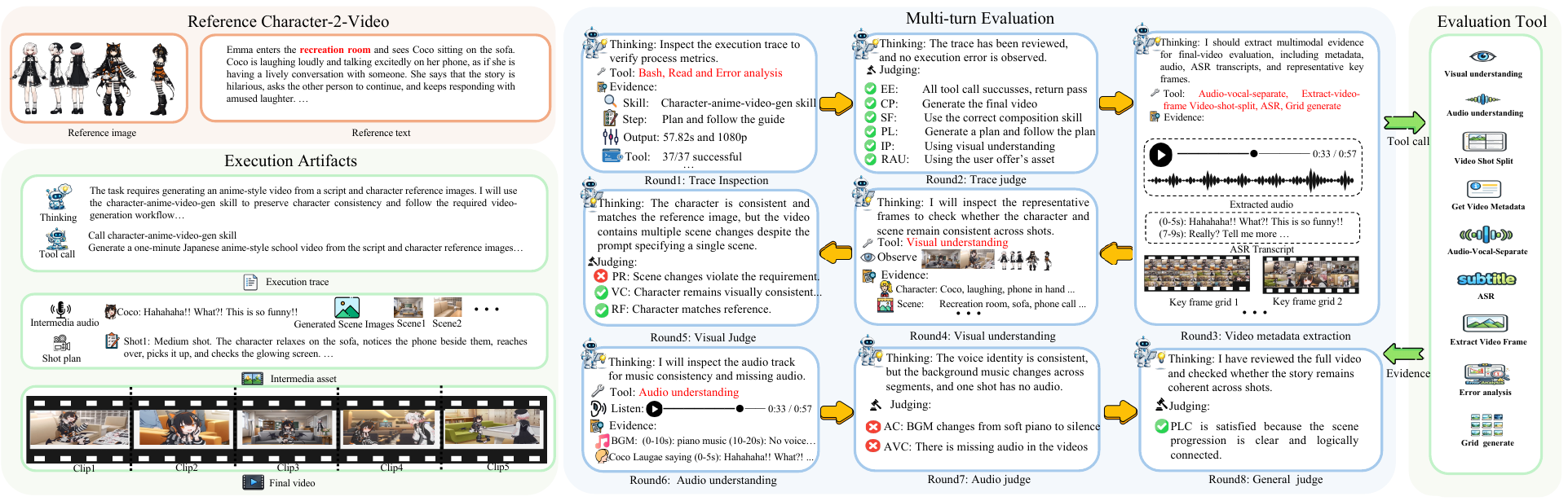}
    \caption{
        Evidence-grounded agent-as-judge evaluation framework. Given the user request, reference assets, execution trace, intermediate artifacts, and final video, the evaluation agent conducts an agentic evaluation. At each round, it invokes evaluation skills to inspect traces, extract video metadata, sample key frames, separate audio, run ASR, analyze errors, and understand visual or audio content. The collected evidence is then used to score process and output metrics and provide textual feedback.
    }
    \label{fig:evaluation_harness}
\end{figure}

\noindent \textbf{Evaluation Protocol.} \label{sec:evaluation_protocol} We use Claude Code \citep{anthropic-claude-code-2026} + DeepSeek V4 \citep{deepseek-v4-2026}, equipped with evaluation skills as a scalable Agent-as-Judge \citep{zheng2023judging,liu2023geval} for inspecting execution traces, tool-call records, generated files, logs, and final videos. The evaluation framework is shown in~\cref{fig:evaluation_harness}.

Specifically, given an evaluation case with task inputs, execution traces, final video outputs, and metric definitions, the evaluation agent assesses each metric in an evidence-grounded manner. With evaluation skills, the agent can extend its native capabilities and dynamically invoke tools to inspect intermediate artifacts, logs, tool-call records, generated clips, and other metadata. For each metric, the agent returns a score from $\{0, 1\}$, supporting evidence, a confidence level, and textual feedback. These scores and feedback are mainly used as refinement signals for skill evolution. The rationale for using binary output-quality metrics is provided in
\cref{app:binary_metrics}.



For final comparison, annotators jointly rank $n=9$ distinct baseline outputs from \cref{tab:main_results,tab:evolution_results} and assign scores from $n-1$ to $0$, allowing ties; the shared OpenClaw + Seed2.0 Composition Skill baseline is counted once. The \textbf{Ranking Score} of baseline $b$ is
\begin{equation}
    \mathrm{RankingScore}(b) =
    \frac{1}{|\mathcal{X}||\mathcal{A}|}
    \sum_{x \in \mathcal{X}} \sum_{a \in \mathcal{A}} r_{x,a}(b),
\end{equation}
where $r_{x,a}(b)$ denotes the score assigned to baseline $b$ by annotator $a$ for input $x$. Each instance is independently annotated by three annotators from diverse backgrounds, including general users, AI video researchers, and film study experts. \cref{app:ranking_demo} details tie handling and the annotation protocol.

\subsection{Evolution Algorithm}
\label{sec:method_overview}
\begin{figure}[htbp]
    \centering
    \includegraphics[width=\textwidth]{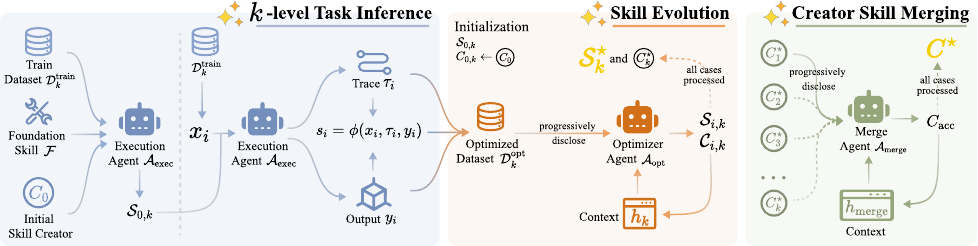}
    \caption{
        Overview of our evolution pipeline. The execution agent $\mathcal{A}_{\mathrm{exec}}$ first performs task inference to generate initial category-level composition skills $\mathcal{S}_{0,k}$ and collect execution traces $\tau$, outputs $y$, and evaluation feedback $s_i$. The optimizer agent $\mathcal{A}_{\mathrm{opt}}$ then progressively refines composition skills and creator skills with feedback to obtain $\mathcal{S}^{\star}_k$ and $C^{\star}_k$. Finally, the merge agent $\mathcal{A}_{\mathrm{merge}}$ integrates multiple creator skills into a single $C^{\star}$.
        }
    \label{fig:evolution_pipeline}
\end{figure}

\cref{fig:evolution_pipeline} illustrates the overall evolution pipeline. The process consists of three stages: task inference, skill evolution, and creator skill merging.

\noindent\textbf{Preliminaries and Objective.} Let $\mathcal{D}^{\mathrm{train}}=\{\mathcal{D}^{\mathrm{train}}_k\}_{k=1}^{K}$ and $\mathcal{D}^{\mathrm{test}}=\{\mathcal{D}^{\mathrm{test}}_k\}_{k=1}^{K}$ denote the train and test splits over $K$ task categories, where $\mathcal{D}^{\mathrm{train}}_k$ and $\mathcal{D}^{\mathrm{test}}_k$ contain cases from category $k$. Let $\mathcal{F}$ be the fixed foundation skill set, $C$ denote a creator skill, $C_0$ the initial creator skill, and $\mathcal{S}_k$ a category-level composition skill. We use three agents: the execution agent $\mathcal{A}_{\mathrm{exec}}$, the optimizer agent $\mathcal{A}_{\mathrm{opt}}$, and the merge agent $\mathcal{A}_{\mathrm{merge}}$. Given an input $x$, an execution trace $\tau$, and an output $y$, the score function $\phi(x,\tau,y)$ is instantiated by the evaluation agent and measures execution and output quality.

We optimize $K$ category-level composition skills using only training cases:
\begin{equation}
    \mathcal{S}^{\star}_k
    =
    \arg\max_{\mathcal{S}_k}
    \sum_{x \in \mathcal{D}^{\mathrm{train}}_k}
    \phi\bigl(x, \mathcal{A}_{\mathrm{exec}}(x, \mathcal{S}_k)\bigr),
\end{equation}
where $\mathcal{A}_{\mathrm{exec}}(x,\mathcal{S}_k)=(\tau,y)$ and $\mathcal{S}^{\star}=\{\mathcal{S}^{\star}_k\}_{k=1}^{K}$. We freeze $\mathcal{S}^{\star}$ before held-out testing and never use test feedback for updates.

We learn category-local creator skills on training categories and merge them for OOD transfer. Each $C_k$ produces
\begin{equation}
    \mathcal{S}_{C_k,k}
    =
    \mathcal{A}_{\mathrm{exec}}(C_k,\mathcal{F},\mathcal{D}^{\mathrm{train}}_k),
\end{equation}
and is optimized using training-case feedback:
\begin{equation}
    C^{\star}_k
    =
    \arg\max_{C_k}
    \sum_{x \in \mathcal{D}^{\mathrm{train}}_k}
    \phi\bigl(x, \mathcal{A}_{\mathrm{exec}}(x, \mathcal{S}_{C_k,k})\bigr).
\end{equation}
The category-local creators are merged into $C^{\star}$, which is frozen before receiving each unseen category's task-level skill query; no OOD execution results or evaluation feedback are used for updates.

\noindent \textbf{Task Inference.}
We first run a forward inference pass to collect optimization data. For each task category $k$, the execution agent uses $C_0$ and $\mathcal{F}$ to produce an initial composition skill,
$\mathcal{S}_{0,k}=\mathcal{A}_{\mathrm{exec}}(C_0,\mathcal{F},\mathcal{D}^{\mathrm{train}}_k)$.
Each train case $x_i \in \mathcal{D}^{\mathrm{train}}_k$ is then executed with $\mathcal{S}_{0,k}$ to obtain
$(\tau_i,y_i)=\mathcal{A}_{\mathrm{exec}}(x_i,\mathcal{S}_{0,k})$,
and scored as $s_i=\phi(x_i,\tau_i,y_i)$. Thus, task inference follows the forward pass:
\begin{equation}
\label{eq:forward_inference}
\begin{aligned}
    C_0,\mathcal{F},\mathcal{D}^{\mathrm{train}}_k
    &\xrightarrow{\mathcal{A}_{\mathrm{exec}}}
    \mathcal{S}_{0,k}, \\
    x_i \in \mathcal{D}^{\mathrm{train}}_k,\ \mathcal{S}_{0,k}
    &\xrightarrow{\mathcal{A}_{\mathrm{exec}}}
    (x_i,\tau_i,y_i)
    \xrightarrow{\phi}
    s_i .
\end{aligned}
\end{equation}
We store $(x_i,\tau_i,y_i,s_i)$ in $\mathcal{D}^{\mathrm{opt}}_k$, yielding $\mathcal{D}^{\mathrm{opt}}=\{\mathcal{D}^{\mathrm{opt}}_k\}_{k=1}^{K}$ and $\mathcal{S}_0=\{\mathcal{S}_{0,k}\}_{k=1}^{K}$.

\noindent \textbf{Skill Evolution.} Using $\mathcal{D}^{\mathrm{opt}}$, $\mathcal{A}_{\mathrm{opt}}$ optimizes each task category in an independent session. For category $k$, we initialize the composition skill as $\mathcal{S}_{0,k}$, initialize a category-local creator skill $C_{0,k}\leftarrow C_0$, and maintain a shared context $h_k$. Cases in $\mathcal{D}^{\mathrm{opt}}_k$ are \textbf{\emph{progressively disclosed}} to $\mathcal{A}_{\mathrm{opt}}$ within the same context. At step $i$, the optimizer receives
$z_i=(h_k,\{x_i,\tau_i,y_i,s_i,\mathcal{S}_{i-1,k},C_{i-1,k}\})$
and updates the current skills to $(\mathcal{S}_{i,k},C_{i,k})$. $\mathcal{A}_{\mathrm{opt}}$ \textbf{\emph{backpropagates}} feedback along the forward chain in \cref{eq:forward_inference}: it analyzes the score, trace, and output to localize failures, first revises the composition skill, and then optimizes the creator skill. The persistent context $h_k$ lets the optimizer accumulate experience across cases and summarize shared patterns within the category. After all cases are processed, we archive the final composition skill set as $\mathcal{S}^{\star}=\{\mathcal{S}^{\star}_k\}_{k=1}^{K}$ and obtain $K$ category-level creator skills $\{C^{\star}_k\}_{k=1}^{K}$.

\noindent \textbf{Creator Skill Merging.}
To integrate experience from multiple category-level creator skills, we merge them sequentially in a shared context $h_{\mathrm{merge}}$. Starting from an accumulator $C_{\mathrm{acc}}\leftarrow C^{\star}_1$, each remaining creator skill $C^{\star}_k$ for $k=2,\ldots,K$ is \textbf{\emph{progressively disclosed}} to $\mathcal{A}_{\mathrm{merge}}$ through $z_k=(h_{\mathrm{merge}},\{C_{\mathrm{acc}},C^{\star}_k\})$, and the accumulator is updated as $C_{\mathrm{acc}}\leftarrow \mathcal{A}_{\mathrm{merge}}(z_k)$. The final accumulator is returned as $C^\star$, a single creator skill that absorbs experience across categories.


\section{Experiments}
\subsection{Baselines}
\label{sec:baselines}
For the execution agent, we evaluate multiple harness environments, including Claude Code \citep{anthropic-claude-code-2026}, Codex \citep{openai-codex-2026}, and the open-source OpenClaw framework \citep{openclaw-2026}, paired with multiple foundation models, including GPT-5.5 \citep{openai-gpt55-2026}, DeepSeek V4 \citep{deepseek-v4-2026}, Opus 4.7 \citep{anthropic-opus47-2026}, and Seed2.0 \citep{bytedance-seed2-2026}. Unless otherwise specified, skill evolution uses OpenClaw with Seed2.0 as the execution agent, Codex with GPT-5.5 as the optimizer agent, and Claude Code with DeepSeek V4 as the evaluation agent to control cost and improve reproducibility. Using three distinct harness-model pairs further helps mitigate potential self-preference bias~\citep{panickssery2024llm}. We compare the following baselines:

\begin{itemize}[leftmargin=*, itemsep=1pt, topsep=2pt, parsep=0pt]
    \item \textbf{No-Composition-Skill:} the agent is provided only with foundation skills and must orchestrate the workflow during execution.
    \item \textbf{Composition-Skill:} the agent first creates a task-specific composition skill that specifies the orchestration procedure over foundation skills, and then executes the task by following it.
    \item \textbf{Self-Evolution:} The agent updates its skills from its own execution outcomes without agent-as-judge feedback.
    \item \textbf{Evolution with Feedback:} Skills are updated using execution results and evidence-grounded feedback produced by the agent-as-judge.
    \item \textbf{Expert Skill:} A composition skill manually constructed by experts.
\end{itemize}

This comparison allows us to assess the effect of externalizing orchestration into composition skills, optimizing them through evolution, and incorporating judge feedback during the evolution process.

\begin{table}[htbp]
\begin{minipage}{\linewidth}
\caption{
Composition-skill results across harness--model pairs on the test set. \textbf{Bold}/\underline{underlined} values are best/second-best before rounding.
}
\label{tab:main_results}
\centering
\small
\setlength{\tabcolsep}{3.0pt}
\renewcommand{\arraystretch}{1.08}
\resizebox{\textwidth}{!}{
\begin{tabular}{lccccccc cccccccc c}
\toprule
\multirow{2}{*}{\textbf{Baseline}}
& \multicolumn{7}{c}{\textbf{Process Metrics}}
& \multicolumn{8}{c}{\textbf{Output Metrics}}
& \multirow{2}{*}{\makecell{\textbf{Ranking}\\\textbf{Score}}} \\
\cmidrule(lr){2-8} \cmidrule(lr){9-16}
& EE & CM & IP & PL & SF & RAU & Avg
& FR & PR & VC & RF & AC & AVC & PLC & Avg
& \\
\midrule

OpenClaw + Seed2.0        & \textbf{0.975} & \textbf{0.986} & 0.886 & 0.869 & \underline{0.921} & 0.895 & 0.922 & 0.728 & 0.255 & 0.493 & \underline{0.582} & \underline{0.567} & 0.605 & 0.631 & 0.552 & 2.96 \\
Claude Code + Seed2.0     & 0.858 & 0.971 & 0.724 & 0.478 & 0.625 & 0.493 & 0.692 & 0.715 & \underline{0.439} & \textbf{0.550} & 0.568 & 0.507 & 0.669 & 0.507 & 0.565 & 3.41 \\
Claude Code + DeepSeek V4 & 0.944 & 0.840 & 0.816 & 0.785 & 0.857 & 0.826 & 0.845 & 0.836 & 0.301 & 0.446 & 0.561 & 0.523 & 0.754 & 0.720 & 0.591 & 4.45 \\
Claude Code + Opus 4.7    & \underline{0.975} & \underline{0.986} & \textbf{1.000} & \underline{0.895} & \textbf{0.966} & \underline{0.961} & \textbf{0.964} & \underline{0.912} & 0.406 & 0.381 & 0.553 & 0.485 & \underline{0.784} & \underline{0.763} & \underline{0.612} & \underline{4.64} \\
Codex + GPT-5.5           & \textbf{0.975} & 0.985 & \underline{0.987} & \textbf{0.914} & 0.911 & \textbf{0.970} & \underline{0.957} & \textbf{0.984} & \textbf{0.458} & \underline{0.521} & \textbf{0.605} & \textbf{0.626} & \textbf{0.840} & \textbf{0.803} & \textbf{0.691} & \textbf{4.90} \\
\bottomrule
\end{tabular}
}
\end{minipage}
\par\vspace{4pt}
\begin{minipage}{\linewidth}
\caption{
Skill-evolution results with OpenClaw + Seed2.0 on the test set. \textbf{Bold}/\underline{underlined} values are best/second-best before rounding.
}
\label{tab:evolution_results}
\centering
\small
\setlength{\tabcolsep}{3.0pt}
\renewcommand{\arraystretch}{1.08}
\resizebox{\textwidth}{!}{
\begin{tabular}{lccccccc cccccccc c}
\toprule
\multirow{2}{*}{\textbf{Baseline}}
& \multicolumn{7}{c}{\textbf{Process Metrics}}
& \multicolumn{8}{c}{\textbf{Output Metrics}}
& \multirow{2}{*}{\makecell{\textbf{Ranking}\\\textbf{Score}}} \\
\cmidrule(lr){2-8} \cmidrule(lr){9-16}
& EE & CM & IP & PL & SF & RAU & Avg
& FR & PR & VC & RF & AC & AVC & PLC & Avg
& \\
\midrule
No Composition Skill      & 0.975 & \textbf{0.992} & 0.571 & 0.465 & --    & 0.820 & 0.765 & 0.789 & 0.254 & 0.395 & 0.565 & 0.482 & \underline{0.772} & 0.562 & 0.546 & 2.82 \\
Composition Skill         & 0.975 & 0.986 & 0.886 & 0.869 & 0.921 & 0.895 & 0.922 & 0.728 & 0.255 & \underline{0.493} & \textbf{0.582} & \textbf{0.567} & 0.605 & 0.631 & 0.552 & 2.96 \\
Self-Evolution            & \textbf{0.986} & \underline{0.986} & \textbf{0.996} & \textbf{0.989} & \textbf{0.988} & 0.982 & \textbf{0.988} & 0.905 & 0.301 & 0.444 & \underline{0.581} & 0.398 & 0.692 & 0.649 & 0.567 & 4.59 \\
Evolution with Feedback   & \underline{0.986} & \textbf{0.992} & \underline{0.971} & \underline{0.985} & \underline{0.985} & \textbf{0.991} & \underline{0.985} & \underline{0.913} & \textbf{0.342} & 0.473 & 0.545 & 0.448 & 0.701 & \underline{0.655} & \underline{0.582} & \underline{5.23} \\
Expert Composition Skill  & 0.965 & 0.971 & 0.948 & 0.985 & 0.964 & \underline{0.986} & 0.970 & \textbf{0.928} & \underline{0.319} & \textbf{0.498} & 0.528 & \underline{0.524} & \textbf{0.799} & \textbf{0.744} & \textbf{0.620} & \textbf{5.63} \\
\bottomrule
\end{tabular}
}
\end{minipage}
\end{table}

\subsection{Main Results}
\label{sec:main_results}

We note that process scores are generally high. This is because they measure whether key execution checkpoints are completed, rather than how well each step is performed. For example, a method receives credit for performing planning if the corresponding action is observed in the trace, while the quality of the resulting plan is primarily reflected in the output metrics.

\noindent \textbf{Harness and Model Comparison: }\label{sec:harness_model_comparison}\cref{tab:main_results} shows that performance varies substantially across different harness-model pairs. Industrial harness settings, such as Codex + GPT-5.5 and Claude Code + Opus 4.7, achieve the strongest overall performance on both process and output metrics.

The results also show clear model-harness compatibility effects. Within Claude Code, replacing Opus 4.7 with Seed2.0 leads to notable drops in both process and output performance, suggesting that different models adapt differently to the same harness interface and orchestration patterns.

\noindent \textbf{Skill Evolution Results: }\label{sec:skill_evolution_results}~\cref{tab:evolution_results} evaluates the optimized composition skills and other variants on test cases under the same OpenClaw + Seed2.0 execution setting. Expert Composition Skill achieves the best output average, showing that human-designed orchestration still provides the strongest reference point. Among automatic methods, Evolution with Feedback achieves a higher output average than Self-Evolution, although its process average is slightly lower. 
This suggests that self-evolution captures many trace-visible procedural
fixes, while agent-as-judge feedback supplies output-level signals that are
difficult for the agent to infer from its own execution alone.
We further evaluate robustness to different case-disclosure and skill merging orders in \cref{app:order_effects}; the average output improvement is also observed over three random orders.

The comparison between No Composition Skill and Composition Skill indicates that foundation skills alone are insufficient for long video generation. No Composition Skill baseline achieves a high Clip Merging score, suggesting that the agent can satisfy the basic target video duration. However, its substantially lower Input Processing and Planning scores indicate that it often omits preparatory steps essential for producing high-quality, cross-clip coherent videos, such as interpreting references, structuring the narrative, and coordinating dependencies across clips. The improvements from Composition Skill therefore demonstrate the importance of explicit procedural guidance for orchestration beyond basic tool use.

\par\noindent\begin{minipage}{\textwidth}
\raggedbottom
\subsection{Generalization}
\label{sec:generalization}

\begin{wraptable}{r}{0.565\textwidth}
\vspace{-6pt}
\captionsetup{skip=4pt}
\caption{
Pairwise generalization results (mean over three independent runs). Test rows compare optimized with initial composition skills; OOD rows compare videos produced using new composition skills generated by optimized versus initial creator skills. Averages are sample-level.
}
\label{tab:generalization_results}
\centering
\footnotesize
\setlength{\tabcolsep}{2pt}
\renewcommand{\arraystretch}{0.90}
\begin{tabular}{lccc}
\toprule
\textbf{Category} & \textbf{Win} & \textbf{Tie} & \textbf{Loss} \\
\specialrule{\lightrulewidth}{0pt}{0pt}
\rowcolor{gray!12}
\multicolumn{4}{l}{\textit{Test Categories}} \\
AI Avatar Video & 30.0\% & 30.0\% & \textbf{40.0\%} \\
One-Sentence Anime Video & \textbf{22.2\%} & 66.7\% & 11.1\% \\
Product Unboxing Video & 12.5\% & 68.8\% & \textbf{18.8\%} \\
SSL Class & \textbf{40.0\%} & 30.0\% & 30.0\% \\
Anime Video & \textbf{40.0\%} & 40.0\% & 20.0\% \\
Character Anime Video & \textbf{16.7\%} & 83.3\% & 0.0\% \\
Object Evolution Video & \textbf{28.6\%} & 71.4\% & 0.0\% \\
Product Video Demo & \textbf{40.0\%} & 60.0\% & 0.0\% \\
Cinematic Video & \textbf{50.0\%} & 30.0\% & 20.0\% \\
Ancient Poetry Teaching & \textbf{30.0\%} & 70.0\% & 0.0\% \\
Audio-Driven Story & 11.1\% & 66.7\% & \textbf{22.2\%} \\
E-Commerce Subject Replace & 8.3\% & 66.7\% & \textbf{25.0\%} \\
Long Video Edit & 12.5\% & 62.5\% & \textbf{25.0\%} \\
\midrule
Test Avg. & \textbf{25.3\%} & 58.2\% & 16.5\% \\
\specialrule{\lightrulewidth}{0pt}{0pt}
\rowcolor{gray!12}
\multicolumn{4}{l}{\textit{OOD Categories}} \\
AI Avatar Tech Content Video & \textbf{37.5\%} & 50.0\% & 12.5\% \\
Theme Video & \textbf{21.1\%} & 68.4\% & 10.5\% \\
Brand Promotion Video & 20.0\% & 53.3\% & \textbf{26.7\%} \\
\midrule
OOD Avg. & \textbf{23.8\%} & 59.7\% & 16.5\% \\
\midrule
All & \textbf{24.9\%} & 58.6\% & 16.5\% \\
\bottomrule
\end{tabular}
\vspace{-6pt}
\end{wraptable}

The OOD results in \cref{tab:generalization_results} evaluate the creator skills by using them to generate new composition skills, which are then executed on OOD cases.

We report means over three independent runs and omit standard deviations due to space constraints.

\cref{tab:generalization_results} shows that the optimized skills (with agent feedback) generalize beyond the training cases. On test categories, optimized composition skills achieve more wins than losses over their initial versions (25.3\% vs. 16.5\%), suggesting reusable orchestration improvements on held-out cases. On OOD categories, videos produced using skills generated by the optimized creator achieve more wins than losses against those produced using skills generated by the initial creator (23.8\% vs. 16.5\%), suggesting transfer of learned orchestration experience to unseen categories. The relatively high tie rates may be attributed to the fact that, for many cases, the initial skills already yield competitive outputs, while pairwise comparison captures only perceptible differences in the generated videos.

\finishresultwrap
\end{minipage}
\par\medskip



\par\noindent\begin{minipage}{\textwidth}
\raggedbottom
\subsection{Human Alignment and Feedback Utility}
\label{sec:human_alignment}

\begin{wraptable}{r}{0.43\textwidth}
\vspace{-6pt}
\captionsetup{skip=4pt}
\caption{
Alignment between Agent-as-Judge scores and human annotations on 200 sampled cases. 
}
\label{tab:human_alignment}
\centering
\footnotesize
\setlength{\tabcolsep}{3pt}
\renewcommand{\arraystretch}{0.85}
\begin{tabular}{l l c c}
\toprule
\textbf{Type} & \textbf{Metric} & \textbf{Exact} & \textbf{$\tau_b$} \\
\midrule
\multirow{5}{*}{Process}
& CM  & 0.995 & 0.892 \\
& IP  & 0.917 & 0.763 \\
& PL  & 0.921 & 0.792 \\
& SF  & 0.949 & 0.773 \\
& RAU & 0.970 & 0.845 \\
\midrule
  \multirow{7}{*}{Output}
  & FR  & 0.945 & 0.815 \\
  & PR  & 0.679 & 0.490 \\
  & VC  & 0.678 & 0.563 \\
  & RF  & 0.768 & 0.601 \\
  & AC  & 0.702 & 0.509 \\
  & AVC & 0.804 & 0.540 \\
  & PLC & 0.763 & 0.390 \\
  \bottomrule
\end{tabular}
\vspace{-6pt}
\end{wraptable}

To validate the reliability of Agent-as-Judge, we compare the scores against human annotations on 200 sampled cases. Details of the metric annotation protocol and inter-annotator
agreement are provided in App.~\ref{app:human_agreement}. As shown in~\cref{tab:human_alignment}, we report exact match and Kendall's $\tau_b$. The alignment scores are higher for process metrics and FR mainly because they are easier to evaluate than open-ended output quality: they can be judged consistently from traces, logs, generated files, and video metadata. Thus, we use human rankings for final comparison and use the
agent-as-judge output primarily as optimization
signal. As shown in
\cref{tab:evolution_results}, this signal is effective: Evolution with
Feedback outperforms Self-Evolution in both Output Avg. (0.582 vs.\
0.567) and human Ranking Score (5.23 vs.\ 4.59).

\finishresultwrap
\end{minipage}
\par\medskip

\subsection{Resource Usage Statistics}
\label{sec:resource_usage}

\cref{tab:resource_usage} reports aggregate resource usage statistics for each baseline. To measure efficiency, we report Relative Generation Time (\textbf{RGT}), defined as the ratio of total execution time to the duration of the generated video (e.g., an RGT of 40 means that generating a one-minute video takes 40 minutes). We also report Vision Content RGT (\textbf{VC-RGT}), defined as the ratio of time spent on video/image generation calls to total execution time. In the OpenClaw + Seed2.0 setting, better performance generally comes with higher resource usage: tool calls, LLM tokens, and RGT increase from No Composition Skill to Expert Composition Skill. Meanwhile, VC-RGT decreases from 0.80 to 0.72, suggesting that stronger methods spend more time on planning, reasoning, and coordination rather than waiting for video/image generation calls. We report the monetary costs in Appendix~\ref{app:execution_costs}.

\begin{table}[H]
\caption{
Resource usage statistics for each baseline. RGT denotes Relative Generation Time, and VC-RGT denotes Vision Content Relative Generation Time.
}
\label{tab:resource_usage}
\centering
\small
\setlength{\tabcolsep}{5pt}
\renewcommand{\arraystretch}{1.05}
\resizebox{\textwidth}{!}{
\begin{tabular}{lcccccc}
\toprule
\textbf{Baseline} & \textbf{RGT} & \textbf{VC-RGT} & \textbf{Tool Calls} & \textbf{Video Tokens} & \textbf{Image Tokens} & \textbf{LLM Tokens} \\
\specialrule{\lightrulewidth}{0pt}{0pt}
\rowcolor{gray!12}
\multicolumn{7}{l}{\textit{Composition Skill Baselines with Different Harness+Model}} \\
Claude Code + Seed2.0     & 36.28 & 0.59 & 28.19 & 1.52M & 4.49K  & 1.31M \\
Claude Code + DeepSeek V4 & 63.41 & 0.66 & 61.89 & 1.65M & 31.59K & 10.09M \\
Claude Code + Opus 4.7    & 29.54 & 0.59 & 32.91 & 1.50M & 9.29K  & 4.27M \\
Codex + GPT-5.5           & 54.67 & 0.78 & 86.65 & 1.83M & 27.21K & 4.32M \\
\specialrule{\lightrulewidth}{0pt}{0pt}
\rowcolor{gray!12}
\multicolumn{7}{l}{\textit{Baselines with OpenClaw + Seed2.0}} \\
No Composition Skill      & 38.29 & 0.80 & 31.30 & 1.66M & 8.02K  & 2.08M \\
Composition Skill         & 48.81 & 0.78 & 37.11 & 1.85M & 3.84K  & 2.72M \\
Self-Evolution            & 61.78 & 0.73 & 46.98 & 1.52M & 10.26K & 3.74M \\
Evolution with Feedback   & 56.08 & 0.74 & 45.95 & 1.72M & 9.17K  & 3.62M \\
Expert Composition Skill  & 65.94 & 0.72 & 43.07 & 1.53M & 53.24K & 3.39M \\
\bottomrule
\end{tabular}
}
\end{table}

\section{Conclusion}

We presented VideoWeaver, an agent harness and benchmark agentic long video generation. VideoWeaver includes 16 task
categories and 285 cases, an evidence-grounded agent-as-judge that evaluates
execution traces and final videos, and a evolution algorithm that distills
feedback into reusable composition and creator skills. Experiments show that
composition skills improve the generation process, while skill evolution
improves output quality and generalizes to unseen cases.

\section*{AI use statement}

We used an AI assistant, Codex, to support code development, debugging, and writing review during the preparation of this work. All AI-assisted outputs were carefully inspected and revised by the authors. The assistant was not used to make autonomous research decisions or alter experimental outcomes. We manually verified all code, data processing steps, experimental results, and paper content to avoid hallucination, bias, or unsupported claims.

\section*{Reproducibility Statement}

To support reproducibility, we provide the benchmark construction and filtering procedure in \cref{sec:task_definition,app:task_collection}, the metric definitions and evaluation protocol in \cref{sec:metrics,sec:evaluation_protocol}, and the skill evolution procedure in \cref{sec:method_overview,alg:skill_evolution}. The harness--model selections are reported in \cref{sec:baselines}, while the foundation-tool configurations and exact model identifiers are provided in \cref{app:foundation_skills}. We further report the human annotation protocol in \cref{app:ranking_demo}, robustness results across random orders in \cref{app:order_effects}, and benchmark execution costs in \cref{app:execution_costs}.

%

\bibliographystyle{iclr2027_conference}
\bibliography{custom}

@misc{videoworldsimulators2024,
  title = {Video generation models as world simulators},
  author = {Tim Brooks and Bill Peebles and Connor Holmes and Will DePue and Yufei Guo and Li Jing and David Schnurr and Joe Taylor and Troy Luhman and Eric Luhman and Clarence Ng and Ricky Wang and Aditya Ramesh},
  year = {2024},
  url = {https://openai.com/research/video-generation-models-as-world-simulators}
}

@misc{google_veo_2024,
  title = {Google I/O 2024: Introducing Veo and Imagen 3 generative AI tools},
  author = {{Google}},
  year = {2024},
  month = {May},
  url = {https://blog.google/technology/ai/google-generative-ai-veo-imagen-3/}
}

@inproceedings{streamingt2v,
  title={StreamingT2V: Consistent, Dynamic, and Extendable Long Video Generation from Text},
  author={Henschel, Roberto and Khachatryan, Levon and Poghosyan, Hayk and Hayrapetyan, Daniil and Tadevosyan, Vahram and Wang, Zhangyang and Navasardyan, Shant and Shi, Humphrey},
  booktitle={Proceedings of the IEEE/CVF Conference on Computer Vision and Pattern Recognition},
  year={2025}
}

@misc{Personavlog,
      title={PersonaVlog: Personalized Multimodal Vlog Generation with Multi-Agent Collaboration and Iterative Self-Correction}, 
      author={Xiaolu Hou and Bing Ma and Jiaxiang Cheng and Xuhua Ren and Kai Yu and Wenyue Li and Tianxiang Zheng and Qinglin Lu},
      year={2025},
      eprint={2508.13602},
      archivePrefix={arXiv},
      primaryClass={cs.CV},
      url={https://arxiv.org/abs/2508.13602}, 
}

@article{Automv,
  title={AutoMV: An Automatic Multi-Agent System for Music Video Generation},
  author={Tang, X. and Lei, X. and Zhu, C. and Chen, S. and Yuan, R. and Li, Y. and Oh, C. and Zhang, G. and Huang, W. and Benetos, E. and others},
  journal={arXiv preprint arXiv:2512.12196},
  year={2025}
}

@misc{script,
      title={The Script is All You Need: An Agentic Framework for Long-Horizon Dialogue-to-Cinematic Video Generation}, 
      author={Chenyu Mu and Xin He and Qu Yang and Wanshun Chen and Jiadi Yao and Huang Liu and Zihao Yi and Bo Zhao and Xingyu Chen and Ruotian Ma and Fanghua Ye and Erkun Yang and Cheng Deng and Zhaopeng Tu and Xiaolong Li and Linus},
      year={2026},
      eprint={2601.17737},
      archivePrefix={arXiv},
      primaryClass={cs.CV},
      url={https://arxiv.org/abs/2601.17737}, 
}

@article{Hollywood,
  title={Hollywood Town: Long-Video Generation via Cross-Modal Multi-Agent Orchestration},
  author={Wei, Z. and Li, M. and Zhang, Z. and Yuan, R. and Hui, P. and Qu, H. and Evans, J. and Agrawala, M. and Rao, A.},
  journal={arXiv preprint arXiv:2510.22431},
  year={2025}
}

@inproceedings{Animaker,
  title={Animaker: Multi-Agent Animated Storytelling with MCTS-Driven Clip Generation},
  author={Shi, H. and Li, Y. and Chen, X. and Wang, L. and Hu, B. and Zhang, M.},
  booktitle={Proceedings of the SIGGRAPH Asia 2025 Conference Papers},
  pages={1--11},
  year={2025}
}

@misc{univa,
title={UniVA: Universal Video Agent towards Open-Source Next-Generation Video Generalist},
author={Zhengyang Liang and Daoan Zhang and Huichi Zhou and Rui Huang and Bobo Li and Yuechen Zhang and Shengqiong Wu and Xiaohan Wang and Jiebo Luo and Lizi Liao and Hao Fei},
year={2025},
eprint={2511.08521},
archivePrefix={arXiv},
primaryClass={cs.CV},
url={https://arxiv.org/abs/2511.08521},
}

@misc{storyagent,
title={StoryAgent: Customized Storytelling Video Generation via Multi-Agent Collaboration},
author={Panwen Hu and Jin Jiang and Jianqi Chen and Mingfei Han and Shengcai Liao and Xiaojun Chang and Xiaodan Liang},
year={2024},
eprint={2411.04925},
archivePrefix={arXiv},
primaryClass={cs.CV},
url={https://arxiv.org/abs/2411.04925},
}

@misc{codirector,
title={Co-Director: Agentic Generative Video Storytelling},
author={Yale Song and Yiwen Song and Nick Losier and Nathan Hodson and Ye Jin and Rhyard Zhu and Yan Xu and Daniel Vlasic and Carina Claassen and Jasmine Leon and Khanh G. LeViet and Zack Chomyn and Joe Timmons and Brett Slatkin and Scott Penberthy and Tomas Pfister},
year={2026},
eprint={2604.24842},
archivePrefix={arXiv},
primaryClass={cs.AI},
url={https://arxiv.org/abs/2604.24842},
}

@article{helios,
  title={Helios: Real Real-Time Long Video Generation Model},
  author={Yuan, Shenghai and Yin, Yuanyang and Li, Zongjian and Huang, Xinwei and Yang, Xiao and Yuan, Li},
  journal={arXiv preprint arXiv:2603.04379},
  year={2026}
}

@article{voyager-2023,
  title={Voyager: An Open-Ended Embodied Agent with Large Language Models},
  author={Wang, Guanzhi and Xie, Yuqi and Jiang, Yunfan and others},
  journal={arXiv preprint arXiv:2305.16291},
  year={2023}
}

@article{autoskill-2026,
  author  = {Yang, Y. and Li, J. and Pan, Q. and others},
  title   = {AutoSkill: Experience-Driven Lifelong Learning via Skill Self-Evolution},
  journal = {Computing Research Repository},
  volume  = {arXiv:2603.01145},
  year    = {2026},
  url     = {https://arxiv.org/abs/2603.01145}
}

@misc{xskill-2026,
      title={XSkill: Continual Learning from Experience and Skills in Multimodal Agents}, 
      author={Guanyu Jiang and Zhaochen Su and Xiaoye Qu and Yi R. Fung},
      year={2026},
      eprint={2603.12056},
      archivePrefix={arXiv},
      primaryClass={cs.AI},
      url={https://arxiv.org/abs/2603.12056}, 
}

@article{skillflow-2026,
  author  = {Zhang, Ziao and Shi, Kou and Huang, Shiting and Nie, Avery and Zeng, Yu and Zhao, Yiming and Fang, Zhen and Su, Qishen and Qiu, Haibo and Yang, Wei and Ren, Qingnan and Zou, Shun and Huang, Wenxuan and Chen, Lin and Chen, Zehui and Zhao, Feng},
  title   = {SkillFlow: Benchmarking Lifelong Skill Discovery and Evolution for Autonomous Agents},
  journal = {Computing Research Repository},
  volume  = {arXiv:2604.17308},
  year    = {2026},
  url     = {https://arxiv.org/abs/2604.17308}
}

@article{continualskillbench-2026,
  author  = {Guan, Tianyi and Wang, Yiding and Yang, Haotong and Cao, Siyuan and Liu, Shirui and Hu, Yi and Li, Jiaqi and Zhang, Muhan},
  title   = {ContinualSkillBench: Can LLM Agents Truly Evolve Their Capabilities?},
  journal = {Computing Research Repository},
  volume  = {arXiv:2608.03874},
  year    = {2026},
  url     = {https://arxiv.org/abs/2608.03874}
}

@misc{coevo-skills-2026,
      title={CoEvoSkills: Self-Evolving Agent Skills via Co-Evolutionary Verification}, 
      author={Hanrong Zhang and Shicheng Fan and Henry Peng Zou and Yankai Chen and Zhenting Wang and Jiayu Zhou and Chengze Li and Wei-Chieh Huang and Yifei Yao and Kening Zheng and Xue and Liu and Xiaoxiao Li and Philip S. Yu},
      year={2026},
      eprint={2604.01687},
      archivePrefix={arXiv},
      primaryClass={cs.AI},
      url={https://arxiv.org/abs/2604.01687}, 
}

@misc{memskill-2026,
      title={MemSkill: Learning and Evolving Memory Skills for Self-Evolving Agents}, 
      author={Haozhen Zhang and Quanyu Long and Jianzhu Bao and Tao Feng and Weizhi Zhang and Haodong Yue and Wenya Wang},
      year={2026},
      eprint={2602.02474},
      archivePrefix={arXiv},
      primaryClass={cs.CL},
      url={https://arxiv.org/abs/2602.02474}, 
}

@misc{evoskill-2026,
      title={EvoSkill: Automated Skill Discovery for Multi-Agent Systems}, 
      author={Salaheddin Alzubi and Noah Provenzano and Jaydon Bingham and Weiyuan Chen and Tu Vu},
      year={2026},
      eprint={2603.02766},
      archivePrefix={arXiv},
      primaryClass={cs.AI},
      url={https://arxiv.org/abs/2603.02766}, 
}

@misc{abstral-2026,
      title={ABSTRAL: Automatic Design of Multi-Agent Systems Through Iterative Refinement and Topology Optimization}, 
      author={Weijia Song and Jiashu Yue and Zhe Pang},
      year={2026},
      eprint={2603.22791},
      archivePrefix={arXiv},
      primaryClass={cs.AI},
      url={https://arxiv.org/abs/2603.22791}, 
}

@misc{trace2skill-2026,
      title={Trace2Skill: Distill Trajectory-Local Lessons into Transferable Agent Skills}, 
      author={Jingwei Ni and Yihao Liu and Xinpeng Liu and Yutao Sun and Mengyu Zhou and Pengyu Cheng and Dexin Wang and Erchao Zhao and Xiaoxi Jiang and Guanjun Jiang},
      year={2026},
      eprint={2603.25158},
      archivePrefix={arXiv},
      primaryClass={cs.AI},
      url={https://arxiv.org/abs/2603.25158}, 
}

@misc{memrl-2026,
      title={MemRL: Self-Evolving Agents via Runtime Reinforcement Learning on Episodic Memory}, 
      author={Shengtao Zhang and Jiaqian Wang and Ruiwen Zhou and Junwei Liao and Yuchen Feng and Zhuo Li and Yujie Zheng and Weinan Zhang and Ying Wen and Zhiyu Li and Feiyu Xiong and Yutao Qi and Bo Tang and Muning Wen},
      year={2026},
      eprint={2601.03192},
      archivePrefix={arXiv},
      primaryClass={cs.CL},
      url={https://arxiv.org/abs/2601.03192}, 
}

@inproceedings{skillforge-2026,
   title={SkillForge: Forging Domain-Specific, Self-Evolving Agent Skills in Cloud Technical Support},
   url={http://dx.doi.org/10.1145/3805712.3808466},
   DOI={10.1145/3805712.3808466},
   booktitle={Proceedings of the 49th International ACM SIGIR Conference on Research and Development in Information Retrieval},
   publisher={ACM},
   author={Liu, Xingyan and Luo, Xiyue and Li, Linyu and Huang, Ganghong and Liu, Jianfeng and Qiao, Honglin},
   year={2026},
   month=jul, pages={4763--4768} }

@misc{medical-imaging-agents-2026,
      title={Evolving Medical Imaging Agents via Experience-driven Self-skill Discovery}, 
      author={Lin Fan and Pengyu Dai and Zhipeng Deng and Haolin Wang and Xun Gong and Yefeng Zheng and Yafei Ou},
      year={2026},
      eprint={2603.05860},
      archivePrefix={arXiv},
      primaryClass={cs.AI},
      url={https://arxiv.org/abs/2603.05860}, 
}

@misc{memento-skills-2026,
      title={Memento-Skills: Let Agents Design Agents}, 
      author={Huichi Zhou and Siyuan Guo and Anjie Liu and Zhongwei Yu and Ziqin Gong and Bowen Zhao and Zhixun Chen and Menglong Zhang and Yihang Chen and Jinsong Li and Runyu Yang and Qiangbin Liu and Xinlei Yu and Jianmin Zhou and Na Wang and Chunyang Sun and Jun Wang},
      year={2026},
      eprint={2603.18743},
      archivePrefix={arXiv},
      primaryClass={cs.AI},
      url={https://arxiv.org/abs/2603.18743}, 
}

@misc{metaclaw-2026,
      title={MetaClaw: Just Talk -- An Agent That Meta-Learns and Evolves in the Wild}, 
      author={Peng Xia and Jianwen Chen and Xinyu Yang and Haoqin Tu and Jiaqi Liu and Kaiwen Xiong and Siwei Han and Shi Qiu and Haonian Ji and Yuyin Zhou and Zeyu Zheng and Cihang Xie and Huaxiu Yao},
      year={2026},
      eprint={2603.17187},
      archivePrefix={arXiv},
      primaryClass={cs.LG},
      url={https://arxiv.org/abs/2603.17187}, 
}

@misc{skalse2025definingcharacterizingrewardhacking,
      title={Defining and Characterizing Reward Hacking}, 
      author={Joar Skalse and Nikolaus H. R. Howe and Dmitrii Krasheninnikov and David Krueger},
      year={2025},
      eprint={2209.13085},
      archivePrefix={arXiv},
      primaryClass={cs.LG},
      url={https://arxiv.org/abs/2209.13085}, 
}

@misc{lightman2023letsverifystepstep,
      title={Let's Verify Step by Step}, 
      author={Hunter Lightman and Vineet Kosaraju and Yura Burda and Harri Edwards and Bowen Baker and Teddy Lee and Jan Leike and John Schulman and Ilya Sutskever and Karl Cobbe},
      year={2023},
      eprint={2305.20050},
      archivePrefix={arXiv},
      primaryClass={cs.LG},
      url={https://arxiv.org/abs/2305.20050}, 
}

@misc{deshpande2025trailtracereasoningagentic,
      title={TRAIL: Trace Reasoning and Agentic Issue Localization}, 
      author={Darshan Deshpande and Varun Gangal and Hersh Mehta and Jitin Krishnan and Anand Kannappan and Rebecca Qian},
      year={2025},
      eprint={2505.08638},
      archivePrefix={arXiv},
      primaryClass={cs.AI},
      url={https://arxiv.org/abs/2505.08638}, 
}

@misc{barke2026agentrxdiagnosingaiagent,
      title={AgentRx: Diagnosing AI Agent Failures from Execution Trajectories}, 
      author={Shraddha Barke and Arnav Goyal and Alind Khare and Avaljot Singh and Suman Nath and Chetan Bansal},
      year={2026},
      eprint={2602.02475},
      archivePrefix={arXiv},
      primaryClass={cs.AI},
      url={https://arxiv.org/abs/2602.02475}, 
}

@article{vbench-2023,
  title={VBench: Comprehensive Benchmark Suite for Video Generative Models},
  author={Huang, Ziqi and He, Yinan and Yu, Jiashuo and Zhang, Fan and Si, Chenyang and Jiang, Yuming and Zhang, Yuanhan and Wu, Tianxing and Jin, Qingyang and Chanpaisit, Nattapol and Wang, Yaohui and Chen, Xinyuan and Wang, Limin and Lin, Dahua and Qiao, Yu and Liu, Ziwei},
  journal={arXiv preprint arXiv:2311.17982},
  year={2023},
  url={https://arxiv.org/abs/2311.17982}
}

@misc{videophy-2024,
      title={VideoPhy: Evaluating Physical Commonsense for Video Generation}, 
      author={Hritik Bansal and Zongyu Lin and Tianyi Xie and Zeshun Zong and Michal Yarom and Yonatan Bitton and Chenfanfu Jiang and Yizhou Sun and Kai-Wei Chang and Aditya Grover},
      year={2024},
      eprint={2406.03520},
      archivePrefix={arXiv},
      primaryClass={cs.CV},
      url={https://arxiv.org/abs/2406.03520}, 
}

@article{visionreward-2024,
  title={VisionReward: Fine-Grained Multi-Dimensional Human Preference Learning for Image and Video Generation},
  author={Xu, Jiazheng and Huang, Yu and Cheng, Jiale and others},
  journal={arXiv preprint arXiv:2412.21059},
  year={2024},
  url={https://arxiv.org/abs/2412.21059}
}

@article{videobench-2025,
  title={Video-Bench: Human-Aligned Video Generation Benchmark},
  author={Han, Hui and Li, Siyuan and Chen, Jiaqi and others},
  journal={arXiv preprint arXiv:2504.04907},
  year={2025},
  url={https://arxiv.org/abs/2504.04907}
}

@article{univbench-2026,
  title={UniVBench: Towards Unified Evaluation for Video Foundation Models},
  author={Wei, Jianhui and Zhang, Xiaotian and Li, Yichen and others},
  journal={arXiv preprint arXiv:2602.21835},
  year={2026},
  url={https://arxiv.org/abs/2602.21835}
}

@article{clvgbench-2026,
  title={How Far Are Video Models from True Multimodal Reasoning?},
  author={Zhang, Xiaotian and Wei, Jianhui and Wang, Yuan and others},
  journal={arXiv preprint arXiv:2604.19193},
  year={2026},
  url={https://arxiv.org/abs/2604.19193}
}

@misc{openai-gpt55-2026,
  title={GPT-5.5 System Card},
  author={{OpenAI}},
  year={2026},
  howpublished={\url{https://openai.com/index/gpt-5-5-system-card/}},
  note={Accessed: 2026-05-24}
}

@misc{bytedance-seed2-2026,
  title={Seed2.0},
  author={{ByteDance Seed}},
  year={2026},
  howpublished={\url{https://github.com/ByteDance-Seed/Seed2.0}},
  note={Accessed: 2026-05-24}
}

@misc{anthropic-claude-code-2026,
  title={Claude Code Overview},
  author={{Anthropic}},
  year={2026},
  howpublished={\url{https://code.claude.com/docs/en/overview}},
  note={Accessed: 2026-05-24}
}

@misc{anthropic-opus47-2026,
  title={Claude Opus 4.7 System Card},
  author={{Anthropic}},
  year={2026},
  howpublished={\url{https://www.anthropic.com/system-cards}},
  note={Accessed: 2026-05-24}
}

@misc{openai-codex-2026,
  title={Introducing the Codex App},
  author={{OpenAI}},
  year={2026},
  howpublished={\url{https://openai.com/index/introducing-the-codex-app/}},
  note={Accessed: 2026-05-24}
}

@misc{deepseek-v4-2026,
  title={DeepSeek-V4},
  author={{DeepSeek-AI}},
  year={2026},
  howpublished={\url{https://www.deepseek.com/en/transparency/}},
  note={Accessed: 2026-05-24}
}

@misc{openclaw-2026,
  title={{OpenClaw}},
  author={{OpenClaw}},
  year={2026},
  howpublished={\url{https://github.com/openclaw/openclaw}},
  note={Accessed: 2026-09-24}
}

@article{liu2023geval,
  title={G-Eval: NLG Evaluation using GPT-4 with Better Human Alignment},
  author={Liu, Yang and Iter, Dan and Xu, Yichong and Wang, Shuohang and Xu, Ruochen and Zhu, Chenguang},
  journal={arXiv preprint arXiv:2303.16634},
  year={2023},
  url={https://arxiv.org/abs/2303.16634}
}

@article{zheng2023judging,
  title={Judging LLM-as-a-Judge with MT-Bench and Chatbot Arena},
  author={Zheng, Lianmin and Chiang, Wei-Lin and Sheng, Ying and Zhuang, Siyuan and Wu, Zhanghao and Zhuang, Yonghao and Lin, Zi and Li, Zhuohan and Li, Dacheng and Xing, Eric P. and Zhang, Hao and Gonzalez, Joseph E. and Stoica, Ion},
  journal={arXiv preprint arXiv:2306.05685},
  year={2023},
  url={https://arxiv.org/abs/2306.05685}
}

@article{panickssery2024llm,
  title={Llm evaluators recognize and favor their own generations},
  author={Panickssery, Arjun and Bowman, Samuel R and Feng, Shi},
  journal={Advances in Neural Information Processing Systems},
  volume={37},
  pages={68772--68802},
  year={2024}
}

@misc{zhang2025storymemmultishotlongvideo,
      title={StoryMem: Multi-shot Long Video Storytelling with Memory}, 
      author={Kaiwen Zhang and Liming Jiang and Angtian Wang and Jacob Zhiyuan Fang and Tiancheng Zhi and Qing Yan and Hao Kang and Xin Lu and Xingang Pan},
      year={2025},
      eprint={2512.19539},
      archivePrefix={arXiv},
      primaryClass={cs.CV},
      url={https://arxiv.org/abs/2512.19539}, 
}

@misc{song2026vqqaagenticapproachvideo,
      title={VQQA: An Agentic Approach for Video Evaluation and Quality Improvement}, 
      author={Yiwen Song and Tomas Pfister and Yale Song},
      year={2026},
      eprint={2603.12310},
      archivePrefix={arXiv},
      primaryClass={cs.CV},
      url={https://arxiv.org/abs/2603.12310}, 
}

@misc{zhang2025evaluationagentefficientpromptable,
      title={Evaluation Agent: Efficient and Promptable Evaluation Framework for Visual Generative Models}, 
      author={Fan Zhang and Shulin Tian and Ziqi Huang and Yu Qiao and Ziwei Liu},
      year={2025},
      eprint={2412.09645},
      archivePrefix={arXiv},
      primaryClass={cs.CV},
      url={https://arxiv.org/abs/2412.09645}, 
}

@misc{cui2026lollongerlongerscaling,
      title={LoL: Longer than Longer, Scaling Video Generation to Hour}, 
      author={Justin Cui and Jie Wu and Ming Li and Tao Yang and Xiaojie Li and Rui Wang and Andrew Bai and Yuanhao Ban and Cho-Jui Hsieh},
      year={2026},
      eprint={2601.16914},
      archivePrefix={arXiv},
      primaryClass={cs.CV},
      url={https://arxiv.org/abs/2601.16914}, 
}

@misc{li2025stablevideoinfinityinfinitelength,
      title={Stable Video Infinity: Infinite-Length Video Generation with Error Recycling}, 
      author={Wuyang Li and Wentao Pan and Po-Chien Luan and Yang Gao and Alexandre Alahi},
      year={2025},
      eprint={2510.09212},
      archivePrefix={arXiv},
      primaryClass={cs.CV},
      url={https://arxiv.org/abs/2510.09212}, 
}

@misc{jiang2025lovicefficientlongvideo,
      title={LoViC: Efficient Long Video Generation with Context Compression}, 
      author={Jiaxiu Jiang and Wenbo Li and Jingjing Ren and Yuping Qiu and Yong Guo and Xiaogang Xu and Han Wu and Wangmeng Zuo},
      year={2025},
      eprint={2507.12952},
      archivePrefix={arXiv},
      primaryClass={cs.CV},
      url={https://arxiv.org/abs/2507.12952}, 
}

@misc{cao2026agenticvbenchaiagentscomplete,
      title={AgenticVBench: Can AI Agents Complete Real-World Post-Production Tasks?}, 
      author={Zongheng Cao and Yi Zheng and Rui Song and Xinyu Hu},
      year={2026},
      eprint={2605.27705},
      archivePrefix={arXiv},
      primaryClass={cs.CR},
      url={https://arxiv.org/abs/2605.27705}, 
}

@misc{li2026skillsbenchbenchmarkingagentskills,
      title={SkillsBench: Benchmarking How Well Agent Skills Work Across Diverse Tasks}, 
      author={Xiangyi Li and Yimin Liu and Wenbo Chen and Bingran You and Zonglin Di and Yifeng He and Shenghan Zheng and Kyoung Whan Choe and Jiankai Sun and Shuyi Wang and Chujun Tao and Binxu Li and Xuandong Zhao and Hejia Geng and Xiaojun Wu and Junwei Zhou and Xiaokun Chen and Hanwen Xing and Yubo Li and Qunhong Zeng and Di Wang and Yuanli Wang and Roey Ben Chaim and Penghao Jiang and Haotian Shen and Luyang Kong and Xinyi Liu and Runhui Wang and Xuanqing Liu and Jiachen Li and Xin Lan and Yueqian Lin and Wengao Ye and Junwei He and Songlin Li and Yue Zhang and Yipeng Gao and Yijiang Li and Ze Ma and Liqiang Jing and Tianyu Wang and Kaixin Li and Yiqi Xue and Haoran Lyu and Yizhuo He and Yuchen Tian and Shutong Wu and Bowei Wang and Yixuan Gao and Bo Chen and Litong Liu and Sikai Cheng and Jiajun Bao and Shuaicheng Tong and Shuwen Xu and Terry Yue Zhuo and Tinghan Ye and Qi Qi and Miao Li and Longtai Liao and Zelin Tan and Chang Shi and Xilin Tang and Srinath Tankasala and Boqin Yuan and Yaoyao Qian and Jianhong Tu and Chenguang Wang and Yizhou Sun and Wei Wang and Aaron Taylor and Ziyue Yang and Changkun Guan and Zhikang Dong and Xinyu Zhang and Steven Dillmann and Han-chung Lee and Dawn Song},
      year={2026},
      eprint={2602.12670},
      archivePrefix={arXiv},
      primaryClass={cs.AI},
      url={https://arxiv.org/abs/2602.12670}, 
}

@misc{yang2026skilloptexecutivestrategyselfevolving,
      title={SkillOpt: Executive Strategy for Self-Evolving Agent Skills}, 
      author={Yifan Yang and Ziyang Gong and Weiquan Huang and Qihao Yang and Ziwei Zhou and Zisu Huang and Yan Li and Xuemei Gao and Qi Dai and Bei Liu and Kai Qiu and Yuqing Yang and Dongdong Chen and Xue Yang and Chong Luo},
      year={2026},
      eprint={2605.23904},
      archivePrefix={arXiv},
      primaryClass={cs.AI},
      url={https://arxiv.org/abs/2605.23904}, 
}

@misc{huang2026vimaxagenticvideogeneration,
      title={ViMax: Agentic Video Generation}, 
      author={Lingxuan Huang and Sizhe He and Hengji Zhou and Liqiang Nie and Lianghao Xia and Chao Huang},
      year={2026},
      eprint={2606.07649},
      archivePrefix={arXiv},
      primaryClass={cs.CV},
      url={https://arxiv.org/abs/2606.07649}, 
}

@misc{li2026frameworkersdynamicmultiagentframework,
      title={FRAMEWORKERS: A Dynamic Multi-Agent Framework for AI-Generated Video Production}, 
      author={Zhendong Li and Lei Sun and Letian Shi and Deheng Zhang and Ruibo Ming and Mengshun Hu and Dannong Xu and Jian Wang and Danda Paudel and Luc Van Gool and Jinjin Gu},
      year={2026},
      eprint={2608.29814},
      archivePrefix={arXiv},
      primaryClass={cs.AI},
      url={https://arxiv.org/abs/2608.29814}, 
}

\appendix
\clearpage

\section{Benchmark Statistics}
\label[appendix]{app:benchmark_statistics}

This section summarizes the scale and evaluation metrics of the benchmarks in
\cref{tab:benchmark_scale}. We count fixed, explicitly named leaf-level
metrics and exclude averages or other scores derived solely by aggregating
those metrics. When a benchmark uses task-specific verifiers or rubric sets
whose contents vary across tasks, we report them as task-specific rather than
assigning a fixed metric count.

\subsection{Video-Generation and Production Benchmarks}

\paragraph{StoryAgent~\citep{storyagent}.}
The public-data evaluation contains 90 samples: five subjects from PororoSV
and four from FlintstonesSV, with ten four-shot samples for each subject. An
additional open-domain set contains eight subjects. The four reference-based
metrics are FVD, PSNR, SSIM, and LPIPS; the five open-domain metrics are
Inception Score, CLIP-score, CLIP-temp, Warping Error, and Flow-score, giving
nine distinct outcome metrics in total. StoryAgent does not separately score
the generation process.

\paragraph{GenAD-Bench~\citep{codirector}.}
GenAD-Bench contains 400 advertising scenarios for 200 fictional products,
with 200 scenarios used for hill-climbing and 200 for validation. It evaluates
Visual Asset Fidelity, Demographic Alignment, Marketing Appeal, and Visual
Quality. These four metrics evaluate the produced advertisements rather than
the intermediate process.

\paragraph{ThemeVlogEval~\citep{Personavlog}.}
The reported source data contain ten celebrity portraits and two visual
styles, together with LLM-generated and manually screened themes whose total
number is not stated. Its twelve outcome metrics comprise four storyboard
metrics (Story Interest, Temporal Continuity, Behavioral Diversity, and
Thematic Consistency), two keyframe metrics (Text--Image Alignment and
Character Consistency), and six video metrics (Subject Consistency,
Background Consistency, Motion Smoothness, Dynamic Degree, Aesthetic Quality,
and Imaging Quality). It does not define process metrics.

\paragraph{AutoMV~\citep{Automv}.}
AutoMV evaluates 30 professionally released songs. Its outcome evaluation
contains ImageBind Score and twelve rubric criteria: Character Consistency,
Physical Authenticity, Lip-sync Accuracy, Visual Style Harmony, Shot
Continuity, Audio--Visual Correlation, Musical Theme Relevance, Storytelling,
Emotional Expression, Visual Composition and Quality, Creativity, and AI
Novelty. We therefore count thirteen outcome metrics and no process metrics.

\paragraph{ST-Bench~\citep{zhang2025storymemmultishotlongvideo}.}
ST-Bench contains 30 long story scripts, each comprising 8--12 shots, and 300
shot-level prompts in total. Its five outcome metrics are Aesthetic Quality,
Global Prompt Following, Single-shot Prompt Following, Overall Cross-shot
Consistency, and Top-10 Cross-shot Consistency. It does not evaluate the
generation process separately.

\paragraph{StreamingT2V~\citep{streamingt2v}.}
StreamingT2V uses 50 prompts and reports three quantitative outcome metrics:
Motion-Aware Warp Error (MAWE), the number of detected scene cuts (SCuts), and
CLIP text--video alignment. It does not report process metrics.

\paragraph{HeliosBench~\citep{helios}.}
HeliosBench contains 240 prompts spanning four duration regimes of 81, 240,
720, and 1,440 frames. Its five base quality metrics are Aesthetic, Dynamic,
Motion Smoothness, Semantic, and Naturalness. It additionally reports drifting
for Aesthetic, Motion Smoothness, Semantic, and Naturalness, giving nine
leaf-level outcome metrics. Composite totals and throughput scores are not
counted, and the benchmark does not score the generation process.

\paragraph{AniEval~\citep{Animaker}.}
AniEval is evaluated on ten stories; its human study covers 90 videos from
nine methods. Its four primary domains contain fourteen fine-grained outcome
metrics. Overall Video Quality contains VQA-A, VQA-T, and MusIQ. Text--Video
Alignment contains Text--Video Consistency, Text--Story Consistency,
Detection-Score, and Count-Score. Video Consistency contains DreamSim, Face
Consistency, Warping Error, and Semantic Consistency. Motion Quality contains
Action Recognition, Action Strength, and Motion AC-Score. AniEval does not
define process metrics.

\paragraph{UniVA-Bench~\citep{univa}.}
The scale of 130 cases in \cref{tab:benchmark_scale} is derived from the
reported task subsets because the paper does not directly state an aggregate
size. The paper reports ten understanding videos, ten Image/Entities2Video
cases, three ten-case Video2Video settings, ten editing cases, ten segmentation
cases, and a 50-instance planning set, in addition to its LongText2Video set.
The three process metrics are weighted Plan Edit Distance (wPED), Dependency
Coverage (DepCov), and Replanning Quality (ReplanQ). The six leaf-level
outcome metrics are CLIP Score, DINO Score, MLLM Preference, Understanding
Accuracy, J-mean, and F-mean. J\&F-mean is excluded because it aggregates
J-mean and F-mean.

\paragraph{AgenticVBench~\citep{cao2026agenticvbenchaiagentscomplete}.}
AgenticVBench contains 100 tasks: 18 Assembly, 18 Repair, 28 Sequencing, and 36
Repurpose tasks. It evaluates submitted artifacts rather than agent
trajectories. Assembly uses a chance-adjusted correct-selection score; Repair
uses task-dependent programmatic repair measurements; Sequencing combines
Normalized Distance, Longest Increasing Subsequence, and Adjacent Fidelity;
and Repurpose uses task-specific binary rubrics covering Format, Visual,
Narrative, Sound, and optional Penalty items. Because the verifier and rubric
contents vary by task family and instance, we denote its outcome metrics as
task-specific in \cref{tab:benchmark_scale}.

\subsection{General Skill Benchmarks and Methods}

\paragraph{SkillsBench~\citep{li2026skillsbenchbenchmarkingagentskills}.}
SkillsBench contains 87 tasks across 8 domains. Its primary outcome metric
is task-macro pass rate, computed from deterministic pass/fail verification.
Absolute and normalized gains are derived comparisons between conditions. It
does not separately score the task-execution process.

\paragraph{SkillOpt~\citep{yang2026skilloptexecutivestrategyselfevolving}.}
The released SkillOpt manifests contain 191 ALFWorld tasks (39 train, 18
validation, and 134 test), 177 LiveMathematicianBench tasks (35, 17, and 125),
and 246 OfficeQA tasks (50, 24, and 172). Each benchmark uses one native
outcome measure: hard success rate for ALFWorld and hard or exact-match
accuracy for LiveMathematicianBench and OfficeQA. SkillOpt uses execution
traces to optimize skills but does not report a separate process score.

\paragraph{SkillFlow~\citep{skillflow-2026}.}
SkillFlow contains 166 tasks organized into 20 families of 8--9 tasks across
five domains. Its outcome metric is task success rate. It also reports five
process, efficiency, or skill-lifecycle statistics: interaction turns,
monetary cost, output tokens, the number of retained skills, and skill reuse
rate.

\paragraph{ContinualSkillBench~\citep{continualskillbench-2026}.}
ContinualSkillBench contains 500 tasks, comprising 100 sequential subtasks in
each of five domains. Its five leaf-level outcome scores are Exact Match, F1,
Numeric, Rubric Judge, and Programmatic evaluation. Raw Reward and Normalized
Reward aggregate these task-level scores and are therefore excluded from the
count. The benchmark analyzes task reuse, core-skill coverage, and generated
skill libraries, but does not directly score the execution workflow of each
task; we therefore mark process evaluation as absent in
\cref{tab:benchmark_scale}.

\subsection{VideoWeaver}

VideoWeaver contains 285 cases across 16 task categories. Its six process
metrics are Execution Error-Free Rate (EE), Clip Merging (CM), Input
Processing (IP), Planning (PL), Skill Following (SF), and Reference Asset
Usage (RAU). Its seven outcome metrics are Format Requirements (FR), Prompt
Requirements (PR), Visual Consistency (VC), Reference Fidelity (RF), Audio
Consistency (AC), Audio--Video Consistency (AVC), and Plot Logic and Coherence
(PLC).

\section{Human Annotation Details}

\subsection{Ranking Annotation Interface}
\label[appendix]{app:ranking_demo}

\begin{figure}[htbp]
    \centering
    \includegraphics[width=\linewidth]{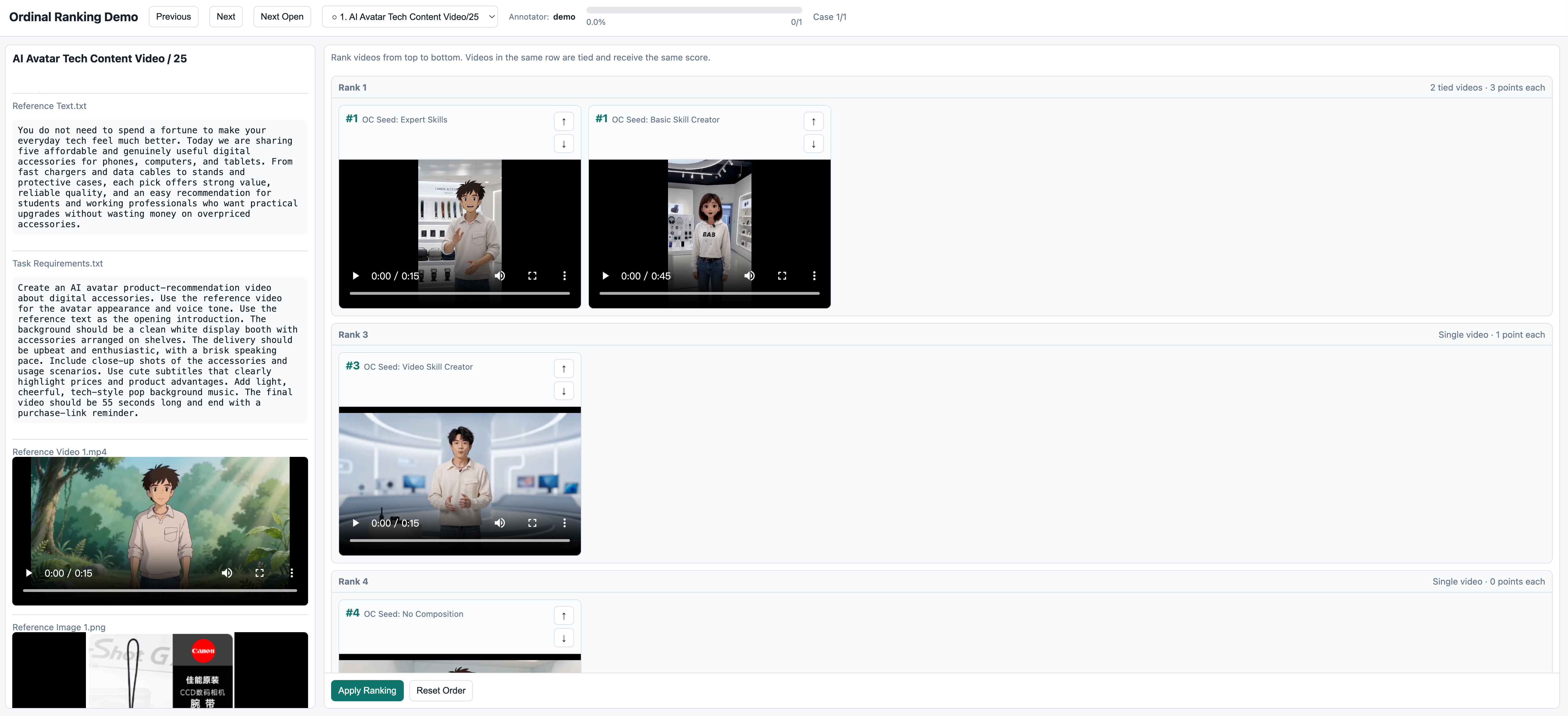}
    \caption{
    Illustration of the ranking annotation interface. Annotators reorder candidate videos for the same input; videos in the same row are tied. Method labels shown in this demonstration are hidden during annotation, and candidate presentation order is randomized.
    }
    \label{fig:ranking_demo}
\end{figure}

We provide the human annotation interface in \cref{fig:ranking_demo}. For each evaluation instance, annotators are shown the user input and candidate videos generated by different baselines, with method identities hidden and candidate presentation order randomized. They rank the videos according to holistic video quality by either dragging videos to reorder them or using the up/down controls in the interface. We use competition ranking: tied videos share the same rank, and subsequent ranks skip the occupied positions (e.g., $1,1,3$). A video at rank $q$ receives $n-q$ points, where $n=9$. Thus, two videos tied for first each receive 8 points, and the next video is ranked third and receives 6 points. Scores are averaged over cases and annotators as in~\cref{sec:evaluation_protocol}.

All annotators were compensated at a rate of 10 USD per hour. The ranking annotation task involves viewing generated videos and task prompts used in our benchmark. It does not require annotators to provide personal information beyond standard annotation-platform account information, and no private or personally identifiable data is included in the released annotations. The collected human judgments are used only for evaluation and analysis in this work, and are not used to train any model.

\subsection{Rationale for Binary Output-Quality Metrics}
\label[appendix]{app:binary_metrics}

The binary output-quality metrics are used to provide scalable diagnostic
feedback for skill evolution, rather than as the sole basis for final
performance comparison. As described in \cref{sec:evaluation_protocol}, our
final cross-system quality comparison is based on human rankings. We choose
binary scores to reduce calibration ambiguity. Fine-grained scales such as
0--5 require defining subtle boundaries between adjacent levels, such as
``slightly coherent,'' ``moderately coherent,'' and ``mostly coherent,''
which may be interpreted inconsistently across evaluators. Evaluators can
more reliably determine whether an obvious audio-video mismatch, reference
mismatch, or plot inconsistency is present than calibrate its precise
degree. We therefore use binary output metrics for diagnostic feedback and
human rankings for final performance comparison.

\subsection{Metric Annotation and Inter-Annotator Agreement}
\label[appendix]{app:human_agreement}

For the human annotations in \cref{tab:human_alignment}, three annotators independently label each evaluation dimension on 200 sampled cases using binary scores in $\{0,1\}$. We measure human--human agreement by strict three-rater unanimity: a case--metric pair receives an agreement indicator of 1 if all three labels are identical, and 0 otherwise, including a two-versus-one split. Let $\mathcal{P}$ denote the set of annotated case--metric pairs and $\ell_{j,p}$ the label from annotator $j$ for pair $p$. The overall Agreement Ratio is
\begin{equation}
    \mathrm{AgreementRatio} = \frac{1}{|\mathcal{P}|}
    \sum_{p\in\mathcal{P}} \mathbf{1}[\ell_{1,p}=\ell_{2,p}=\ell_{3,p}] = 0.67.
\end{equation}
This 67\% measures unanimous agreement among human annotators.

\clearpage
\section{Skill Evolution Algorithm Details}

\begin{algorithm}[H]
\small
\caption{Skill Evolution}
\label{alg:skill_evolution}
\begin{algorithmic}[1]
\Require Train dataset $\mathcal{D}^{\mathrm{train}}=\{\mathcal{D}^{\mathrm{train}}_k\}_{k=1}^{K}$, where each case $i$ contains input $x_i$; fixed foundation skill set $\mathcal{F}$; initial creator skill $C_0$; score function $\phi$; execution agent $\mathcal{A}_{\mathrm{exec}}$; optimizer agent $\mathcal{A}_{\mathrm{opt}}$; merge agent $\mathcal{A}_{\mathrm{merge}}$
\Ensure Optimized composition skills $\mathcal{S}^\star=\{\mathcal{S}^\star_k\}_{k=1}^{K}$ and optimized creator skill $C^\star$

\State $\mathcal{D}^{\mathrm{opt}}, \mathcal{S}_0 \leftarrow \Call{Inference}{\mathcal{D}^{\mathrm{train}}, C_0, \mathcal{F}, \mathcal{A}_{\mathrm{exec}}, \phi}$
\State $\mathcal{S}^\star, \{C^\star_k\}_{k=1}^{K} \leftarrow \Call{Evolve}{\mathcal{D}^{\mathrm{opt}}, \mathcal{S}_0, C_0, \mathcal{A}_{\mathrm{opt}}}$
\State $C^\star \leftarrow \Call{MergeCreators}{\{C^\star_k\}_{k=1}^{K}, \mathcal{A}_{\mathrm{merge}}}$
\State \Return $\mathcal{S}^\star, C^\star$

\Function{Inference}{$\mathcal{D}^{\mathrm{train}}, C_0, \mathcal{F}, \mathcal{A}_{\mathrm{exec}}, \phi$}
    \ForAll{task categories $k \in \{1,\ldots,K\}$ \textbf{in parallel}}
        \State $\mathcal{S}_{0,k} \leftarrow \mathcal{A}_{\mathrm{exec}}(C_0, \mathcal{F}, \mathcal{D}^{\mathrm{train}}_k)$
        \ForAll{cases $i \in \mathcal{D}^{\mathrm{train}}_k$ \textbf{in parallel}}
            \State $(\tau_i, y_i) \leftarrow \mathcal{A}_{\mathrm{exec}}(x_i, \mathcal{S}_{0,k})$
            \State $s_i \leftarrow \phi(x_i, \tau_i, y_i)$
            \State Store $(x_i, \tau_i, y_i, s_i)$ in $\mathcal{D}^{\mathrm{opt}}_k$
        \EndFor
    \EndFor
    \State $\mathcal{D}^{\mathrm{opt}} \leftarrow \{\mathcal{D}^{\mathrm{opt}}_k\}_{k=1}^{K}$, $\mathcal{S}_0 \leftarrow \{\mathcal{S}_{0,k}\}_{k=1}^{K}$
    \State \Return $\mathcal{D}^{\mathrm{opt}}, \mathcal{S}_0$
\EndFunction

\Function{Evolve}{$\mathcal{D}^{\mathrm{opt}}, \mathcal{S}_0, C_0, \mathcal{A}_{\mathrm{opt}}$}
    \ForAll{task categories $k \in \{1,\ldots,K\}$ \textbf{in parallel}}
        \State Initialize $C_{0,k} \leftarrow C_0$ and context $h_k \leftarrow \emptyset$
        \ForAll{cases $i \in \mathcal{D}^{\mathrm{opt}}_k$ \textbf{sequentially}}
            \State $z_i = \left(h_k, \{x_i, \tau_i, y_i, s_i, \mathcal{S}_{i-1,k}, C_{i-1,k}\}\right)$
            \State $(\mathcal{S}_{i,k}, C_{i,k}) \leftarrow \mathcal{A}_{\mathrm{opt}}(z_i)$; $h_k \leftarrow h_k \oplus (\mathcal{S}_{i,k}, C_{i,k})$
        \EndFor
        \State Archive $\mathcal{S}^\star_k \leftarrow \mathcal{S}_{|\mathcal{D}^{\mathrm{opt}}_k|,k}$ and $C^\star_k \leftarrow C_{|\mathcal{D}^{\mathrm{opt}}_k|,k}$
    \EndFor
    \State $\mathcal{S}^\star \leftarrow \{\mathcal{S}^\star_k\}_{k=1}^{K}$
    \State \Return $\mathcal{S}^\star, \{C^\star_k\}_{k=1}^{K}$
\EndFunction

\Function{MergeCreators}{$\{C^\star_k\}_{k=1}^{K}, \mathcal{A}_{\mathrm{merge}}$}
    \State Initialize $C_{\mathrm{acc}} \leftarrow C^\star_1$ and merge context $h_{\mathrm{merge}} \leftarrow \emptyset$
    \For{$k = 2$ to $K$}
        \State $z_k = \left(h_{\mathrm{merge}}, \{C_{\mathrm{acc}}, C^\star_k\}\right)$
        \State $C_{\mathrm{acc}} \leftarrow \mathcal{A}_{\mathrm{merge}}(z_k)$; $h_{\mathrm{merge}} \leftarrow h_{\mathrm{merge}} \oplus C_{\mathrm{acc}}$
    \EndFor
    \State $C^\star \leftarrow C_{\mathrm{acc}}$
    \State \Return $C^\star$
\EndFunction
\end{algorithmic}
\end{algorithm}

\section{Full-Precision Main Results}
\label[appendix]{app:full_precision_results}

\cref{tab:main_results_full,tab:evolution_results_full} report the full-precision values underlying the rounded results in \cref{tab:main_results,tab:evolution_results}. Best and second-best results are determined using these full-precision values.

\begin{table}[H]
\caption{Full-precision results on the test set for the harness and model comparison in \cref{tab:main_results}.}
\label{tab:main_results_full}
\centering
\small
\setlength{\tabcolsep}{3.0pt}
\renewcommand{\arraystretch}{1.08}
\resizebox{\textwidth}{!}{
\begin{tabular}{lccccccc cccccccc c}
\toprule
\multirow{2}{*}{\textbf{Baseline}}
& \multicolumn{7}{c}{\textbf{Process Metrics}}
& \multicolumn{8}{c}{\textbf{Output Metrics}}
& \multirow{2}{*}{\makecell{\textbf{Ranking}\\\textbf{Score}}} \\
\cmidrule(lr){2-8} \cmidrule(lr){9-16}
& EE & CM & IP & PL & SF & RAU & Avg
& FR & PR & VC & RF & AC & AVC & PLC & Avg
& \\
\midrule
OpenClaw + Seed2.0        & \textbf{0.9750} & \textbf{0.9857} & 0.8861 & 0.8691 & \underline{0.9213} & 0.8953 & 0.9221 & 0.7284 & 0.2553 & 0.4929 & \underline{0.5816} & \underline{0.5674} & 0.6050 & 0.6312 & 0.5517 & 2.96 \\
Claude Code + Seed2.0     & 0.8581 & 0.9710 & 0.7237 & 0.4783 & 0.6252 & 0.4928 & 0.6915 & 0.7154 & \underline{0.4388} & \textbf{0.5504} & 0.5683 & 0.5072 & 0.6691 & 0.5072 & 0.5652 & 3.41 \\
Claude Code + DeepSeek V4 & 0.9438 & 0.8404 & 0.8156 & 0.7849 & 0.8567 & 0.8258 & 0.8445 & 0.8355 & 0.3010 & 0.4464 & 0.5606 & 0.5225 & 0.7543 & 0.7197 & 0.5914 & 4.45 \\
Claude Code + Opus 4.7    & \underline{0.9749} & \underline{0.9856} & \textbf{1.0000} & \underline{0.8951} & \textbf{0.9657} & \underline{0.9612} & \textbf{0.9638} & \underline{0.9123} & 0.4055 & 0.3814 & 0.5533 & 0.4845 & \underline{0.7835} & \underline{0.7629} & \underline{0.6119} & \underline{4.64} \\
Codex + GPT-5.5           & \textbf{0.9750} & 0.9849 & \underline{0.9867} & \textbf{0.9137} & 0.9112 & \textbf{0.9698} & \underline{0.9569} & \textbf{0.9843} & \textbf{0.4580} & \underline{0.5210} & \textbf{0.6050} & \textbf{0.6261} & \textbf{0.8403} & \textbf{0.8025} & \textbf{0.6910} & \textbf{4.90} \\
\bottomrule
\end{tabular}
}
\end{table}

\begin{table}[H]
\caption{Full-precision results on the test set for the skill evolution comparison in \cref{tab:evolution_results}.}
\label{tab:evolution_results_full}
\centering
\small
\setlength{\tabcolsep}{3.0pt}
\renewcommand{\arraystretch}{1.08}
\resizebox{\textwidth}{!}{
\begin{tabular}{lccccccc cccccccc c}
\toprule
\multirow{2}{*}{\textbf{Baseline}}
& \multicolumn{7}{c}{\textbf{Process Metrics}}
& \multicolumn{8}{c}{\textbf{Output Metrics}}
& \multirow{2}{*}{\makecell{\textbf{Ranking}\\\textbf{Score}}} \\
\cmidrule(lr){2-8} \cmidrule(lr){9-16}
& EE & CM & IP & PL & SF & RAU & Avg
& FR & PR & VC & RF & AC & AVC & PLC & Avg
& \\
\midrule
No Composition Skill      & 0.9753 & \textbf{0.9919} & 0.5708 & 0.4651 & --     & 0.8195 & 0.7645 & 0.7893 & 0.2536 & 0.3949 & 0.5652 & 0.4819 & \underline{0.7717} & 0.5616 & 0.5455 & 2.82 \\
Composition Skill         & 0.9750 & 0.9857 & 0.8861 & 0.8691 & 0.9213 & 0.8953 & 0.9221 & 0.7284 & 0.2553 & \underline{0.4929} & \textbf{0.5816} & \textbf{0.5674} & 0.6050 & 0.6312 & 0.5517 & 2.96 \\
Self-Evolution            & \textbf{0.9864} & \underline{0.9858} & \textbf{0.9959} & \textbf{0.9890} & \textbf{0.9877} & 0.9823 & \textbf{0.9879} & 0.9047 & 0.3011 & 0.4444 & \underline{0.5806} & 0.3978 & 0.6918 & 0.6487 & 0.5670 & 4.59 \\
Evolution with Feedback   & \underline{0.9860} & \textbf{0.9919} & \underline{0.9707} & \underline{0.9853} & \underline{0.9845} & \textbf{0.9911} & \underline{0.9849} & \underline{0.9128} & \textbf{0.3416} & 0.4733 & 0.5445 & 0.4484 & 0.7011 & \underline{0.6548} & \underline{0.5824} & \underline{5.23} \\
Expert Composition Skill  & 0.9652 & 0.9708 & 0.9481 & 0.9848 & 0.9640 & \underline{0.9860} & 0.9698 & \textbf{0.9283} & \underline{0.3187} & \textbf{0.4982} & 0.5275 & \underline{0.5238} & \textbf{0.7985} & \textbf{0.7436} & \textbf{0.6198} & \textbf{5.63} \\
\bottomrule
\end{tabular}
}
\end{table}

\paragraph{Task completeness and consistency.}
The output metrics in \cref{tab:evolution_results_full} should be interpreted jointly with task completeness. A relevant failure mode is incomplete generation: for example, the initial Composition Skill may produce only two clips totaling 30 seconds for a 60-second request. Such an output fails the duration requirement and may omit requested events or narrative transitions, affecting Format Requirements (FR), Prompt Requirements (PR), and Plot Logic and Coherence (PLC). At the same time, fewer clips and a shorter temporal span provide fewer opportunities for visual, reference, and audio drift, potentially yielding relatively high Visual Consistency (VC), Reference Fidelity (RF), and Audio Consistency (AC) scores on the content that is generated. This offers a possible explanation for the mixed changes from Composition Skill to Evolution with Feedback: FR, PR, and Audio-Video Consistency (AVC) increase by 0.1844, 0.0863, and 0.0961, respectively, and PLC improves more modestly by 0.0236, while VC, RF, and AC decrease.

\section{Order Effect Ablation}
\label[appendix]{app:order_effects}

To address the possible order effect, we conduct additional experiments with different random orders. We use three different random seeds to shuffle both the order of case disclosure and the order of composition-skill merging. The evaluation protocol is kept unchanged.

\begin{table}[H]
\caption{Results under different random orders of case disclosure and composition-skill merging. Values are mean $\pm$ standard deviation over three random seeds. Mean $\Delta$ is the difference between the reported means (Evolution with Feedback minus Self-Evolution).}
\label{tab:order_effects}
\centering
\small
\setlength{\tabcolsep}{6pt}
\renewcommand{\arraystretch}{1.05}
\begin{tabular}{lccc}
\toprule
\textbf{Metric} & \textbf{Evolution with Feedback} & \textbf{Self-Evolution} & \textbf{Mean $\Delta$} \\
\midrule
\multicolumn{4}{l}{\textit{Process Metrics}} \\
EE  & $0.9367 \pm 0.0608$ & $0.9935 \pm 0.0112$ & $-0.0568$ \\
CM  & $0.9465 \pm 0.0927$ & $0.9455 \pm 0.0480$ & $+0.0010$ \\
IP  & $0.9701 \pm 0.0253$ & $0.9529 \pm 0.0510$ & $+0.0172$ \\
PL  & $0.9131 \pm 0.0845$ & $1.0000 \pm 0.0000$ & $-0.0869$ \\
SF  & $0.9959 \pm 0.0070$ & $0.9560 \pm 0.0387$ & $+0.0399$ \\
RAU & $0.9359 \pm 0.0402$ & $0.9089 \pm 0.0508$ & $+0.0270$ \\
Avg & $0.9497 \pm 0.0163$ & $0.9595 \pm 0.0141$ & $-0.0098$ \\
\midrule
\multicolumn{4}{l}{\textit{Output Metrics}} \\
FR  & $0.9216 \pm 0.0558$ & $0.8866 \pm 0.0234$ & $+0.0350$ \\
PR  & $0.2697 \pm 0.0392$ & $0.3511 \pm 0.1251$ & $-0.0814$ \\
VC  & $0.5669 \pm 0.0474$ & $0.5032 \pm 0.1071$ & $+0.0637$ \\
RF  & $0.5640 \pm 0.1255$ & $0.5672 \pm 0.0913$ & $-0.0032$ \\
AC  & $0.5562 \pm 0.0381$ & $0.4499 \pm 0.0720$ & $+0.1063$ \\
AVC & $0.6919 \pm 0.1079$ & $0.6356 \pm 0.1024$ & $+0.0563$ \\
PLC & $0.5796 \pm 0.0361$ & $0.6239 \pm 0.0578$ & $-0.0443$ \\
Avg & $0.5928 \pm 0.0189$ & $0.5739 \pm 0.0422$ & $+0.0189$ \\
\bottomrule
\end{tabular}
\end{table}

Averaged over the three random orders, Evolution with Feedback achieves a higher Output Avg. than Self-Evolution, with a mean difference of $+0.0189$.

\section{Dataset Details}
\label[appendix]{app:dataset_examples}

We construct a benchmark of 16 long-video generation task categories with 285 cases in total. Following the split used in the main experiments, 13 categories are treated as in-distribution categories and split into 119 training cases and 123 held-out test cases, while three categories are held out as OOD categories with 43 cases. Each category is associated with a task-level skill query that describes the intended generation capability and the expected input-output pattern. At 285 cases, VideoWeaver is of the same order of magnitude as existing
skill-related and real-world agentic benchmarks shown in
\cref{tab:benchmark_scale}. Moreover, video generation tasks are generally
costly to construct and evaluate, especially for long-video generation.
Scaling the benchmark much further would potentially increase computation
costs, making the benchmark less accessible and harder to reproduce. We
therefore choose the current scale as a practical balance between task
coverage, evaluation cost, and accessibility.

\subsection{Task Collection and Filtering}
\label[appendix]{app:task_collection}

The user tasks and textual requirements were collected through a platform purposely built for this benchmark. The platform allows users to submit video generation requests with any input modalities, including text, images, videos, audio, and their combinations. Users can also specify constraints such as duration, resolution, aspect ratio, visual style, reference usage, and others. We then filtered the data by keeping approximately one-minute tasks and removing ambiguous requests, grouping similar tasks together, yielding the final benchmark.

\subsection{Dataset Case Examples}
\label[appendix]{app:dataset_cases}

\begin{figure}[htbp]
\centering
\includegraphics[width=0.86\textwidth]{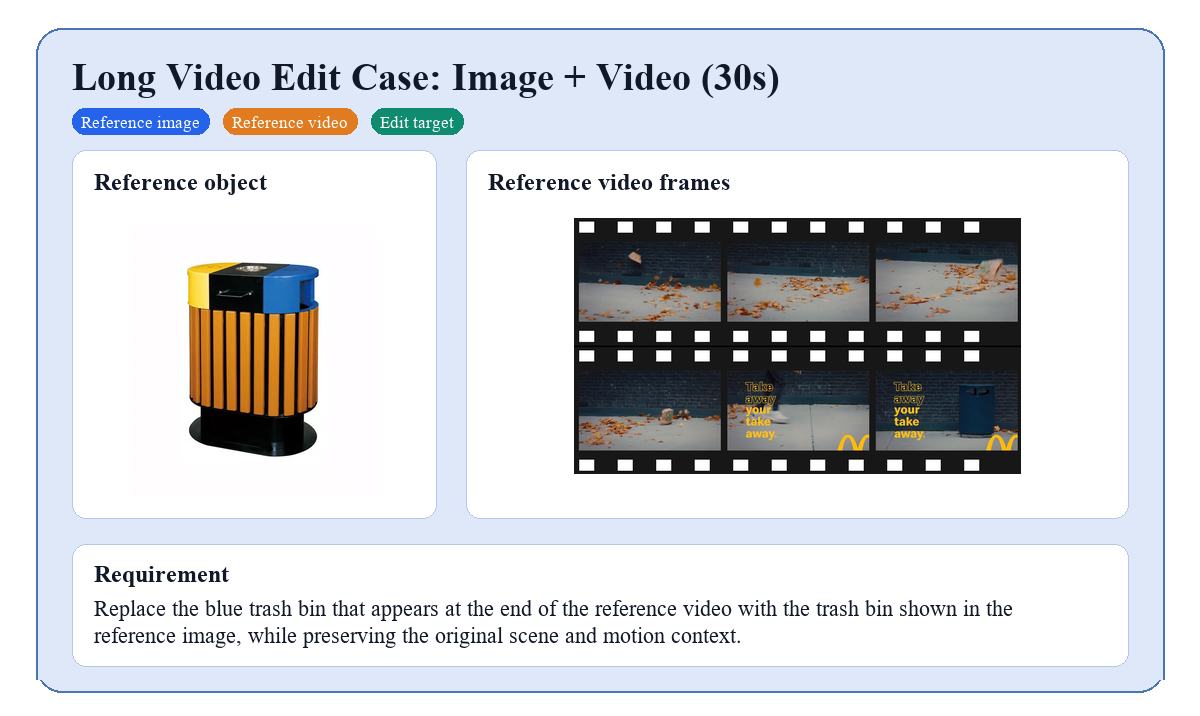}

\includegraphics[width=0.86\textwidth]{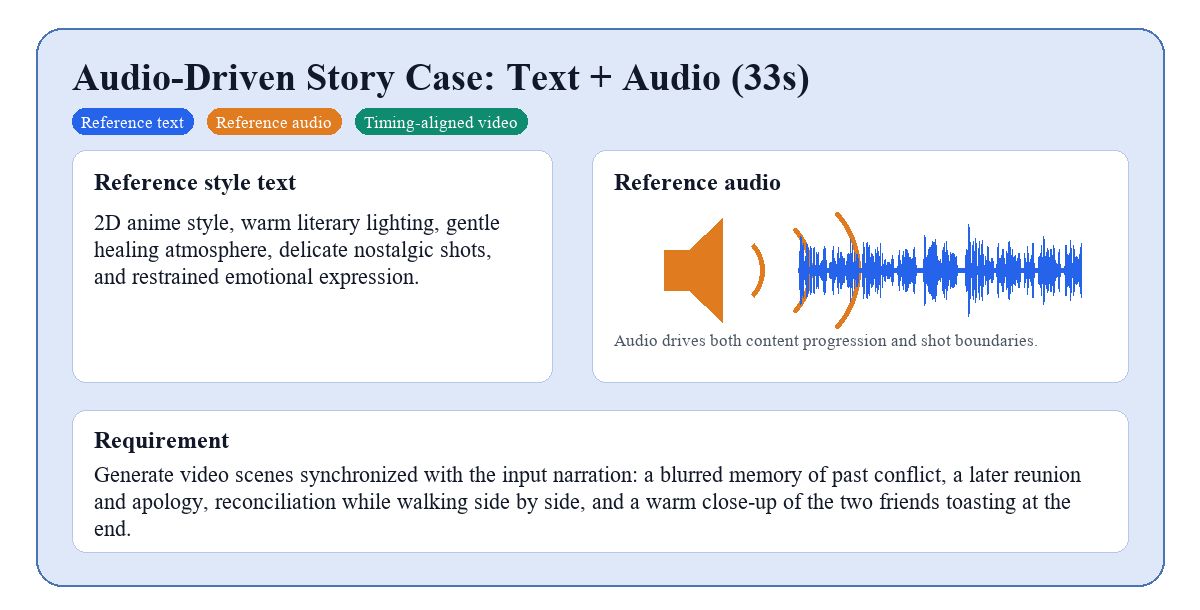}
\caption{Representative dataset cases. Top: a long-video editing case with reference image and reference video inputs. Bottom: an audio-driven story generation case with reference text and reference audio inputs.}
\label{fig:dataset_case_examples}
\end{figure}

\cref{fig:dataset_case_examples} shows two representative cases from the benchmark. The first case illustrates a long-video editing task conditioned on both a reference image and a reference video, where the agent must replace the target object in the final video segment while preserving the original scene context. The second case illustrates an audio-driven story generation task conditioned on a reference text and a reference audio clip, where the generated visual storyline should follow the narration and align scene transitions with the audio.

\clearpage
\subsection{Dataset Skill Queries}

\cref{tab:dataset_skill_queries} lists the dataset categories and their corresponding skill queries. The names and queries are taken from the \texttt{skill\_query.json} file under each dataset directory.

\begin{table}[H]
\caption{Dataset categories and task queries. Categories below the separator are reserved as OOD categories.}
\label{tab:dataset_skill_queries}
\centering
\scriptsize

\begin{tabular}{p{0.24\linewidth}p{0.68\linewidth}}
\toprule
\textbf{Name} & \textbf{Query} \\
\midrule
AI Avatar Video & Generate AI-avatar talking-head videos from narration scripts and avatar reference videos, preserving the speaker appearance, style, voice, target duration, aspect ratio, and other user-specified constraints. \\
One-Sentence Anime Video & Generate anime or animated story videos from a short plot description and user-specified constraints such as duration, aspect ratio, and visual style. \\
Product Unboxing Video & Generate product unboxing videos from product reference images and textual introductions, including product presentation, feature explanation, and close-up demonstrations. \\
SSL Class & Generate children's teaching videos from scripts, teaching requirements, and character references, producing classroom-style instructional clips with coherent teacher-student interactions. \\
Anime Video & Generate anime videos from text and reference anime videos, preserving the reference visual style, tone, color palette, and story requirements. \\
Character Anime Video & Generate character-consistent anime story videos from reference character illustrations and plot text, supporting customized style, scene, duration, and whole-video character consistency. \\
Object Evolution Video & Generate coherent videos showing an object, creature, or scene evolving across time from multi-stage reference images and textual descriptions. \\
Product Video Demo & Generate product demonstration videos from product screenshots and feature descriptions, with narration and visual explanation of key product functions. \\
Cinematic Video & Generate cinematic videos from scripts or long story outlines, following user-specified duration, style, scenes, and visual details. \\
Ancient Poetry Teaching & Generate ancient-poetry teaching animations from poem text and reference images, with narration, pedagogical structure, and coherent visual presentation. \\
Audio-Driven Story & Generate story videos aligned with a provided narration audio file and style references, matching the audio content, timing, and duration. \\
E-Commerce Subject Replace & Replace the product subject or visual style in e-commerce livestream-style short videos while preserving the original selling logic, duration, and rhythm. \\
Long Video Edit & Edit long videos according to text instructions and reference materials, preserving duration and the required identity, scene, object, audio, and background constraints. \\
\midrule
\multicolumn{2}{l}{\textit{OOD Categories}} \\
AI Avatar Tech Content Video & Generate technology-oriented AI-avatar explanation videos from textual requirements, product reference images, and an avatar reference video, preserving the avatar identity, voice, and product-specific content. \\
Theme Video & Generate long-form theme videos from a topic or short text, supporting knowledge explanation, scenery montage, and story narration under user-specified duration and format constraints. \\
Brand Promotion Video & Generate high-quality brand promotion videos from brand references and textual introductions, following user-specified visual style, music, narrative rhythm, and brand consistency requirements. \\
\bottomrule
\end{tabular}
\end{table}

\clearpage
\section{Foundation Skills}
\label[appendix]{app:foundation_skills}

\cref{tab:foundation_skills} lists the foundation skills provided to the agent. These skills expose basic video, image, audio, understanding, metadata, and media-processing capabilities. They are independently invocable and form the primitive action space over which composition skills define longer procedural workflows.

For generation skills, the video generation skill uses ByteDance Doubao Seedance 2.0, with \texttt{doubao-seedance-2-0-fast-260128} as the default model and \texttt{doubao-seedance-2-0-260128} as the higher-quality standard variant. The image generation skill uses Doubao Seedream 5.0 by default (\texttt{doubao-seedream-5-0-260128}) and also supports \texttt{gpt-image-2}. For multimodal perception, the vision-understanding skill uses the Volcengine Ark Doubao multimodal model \texttt{doubao-seed-2-0-pro-260215}, while the audio-understanding skill uses \texttt{doubao-seed-2-0-lite-260428}, which supports native audio input. Speech recognition is implemented with Volcengine BigASR using the \texttt{volc.bigasr.auc\_turbo} resource and \texttt{bigmodel} recognition backend. Text-to-speech uses OpenSpeech TTS V3 with \texttt{seed-tts-2.0} as the default resource. Non-generative media-processing skills, such as metadata extraction, frame extraction, shot splitting, and grid construction, rely on deterministic tools including ffprobe, exiftool, OpenCV, PySceneDetect, and Pillow rather than learned foundation models. We adopt this ByteDance-backed full-modality tool stack consistently across experiments, as it provides unified access to video, image, audio, speech, and multimodal understanding capabilities, thereby simplifying experimental control, reproducibility, and future research extensions. These models are all publicly available through API calls.

\begin{table}[htbp]
\caption{Foundation Skills.}
\label{tab:foundation_skills}
\centering
\scriptsize

\begin{tabular}{p{0.24\linewidth}p{0.68\linewidth}}
\toprule
\textbf{Name} & \textbf{Description} \\
\midrule
add-audio-track & Add an extra audio track to a video while preserving the original audio track. \\
audio-gen & Generate speech audio for narration or dialogue; background music and sound effects are outside this skill's scope. \\
audio-understanding & Analyze audio content with a multimodal audio model, including speaker timbre, BGM style, emotion, continuity, silence, and audio consistency. \\
audio-vocal-separate & Separate vocals from background music or accompaniment in an audio or video file, producing clean vocal and background tracks. \\
automatic-speech-recognition & Transcribe local or online audio/video files into text, JSON, or SRT subtitles. \\
change-fps & Convert a video to a target frame rate. \\
extract-video-frame & Extract a specific frame from a video, such as the first, middle, last, frame-indexed, or timestamped frame. \\
get-image-metadata & Extract image metadata such as resolution, format, file size, and EXIF information. \\
get-output-dir & Return the standardized output directory for the current generation task. \\
get-video-metadata & Extract video metadata such as duration, resolution, frame rate, and frame count. \\
image-gen & Generate or edit images from prompts, reference images, resolution, aspect ratio, model, and seed settings. \\
merge-video & Merge multiple videos after normalizing resolution, aspect ratio, and frame rate. \\
pair-wise-skill-merge & Progressively merge an incoming skill or skill creator into an accumulator skill directory while deduplicating and resolving conflicts. \\
replace-audio & Replace a video's original audio with a specified audio file. \\
resize-image-resolution & Resize an image to a target resolution. \\
resize-video-resolution & Resize a video to a target resolution. \\
split-audio & Separate a video's audio track from its visual stream. \\
split-image & Split an image into a $2\times2$ grid, commonly used to preprocess human reference images for video generation. \\
text-to-speech & Synthesize speech audio from text with configurable voice, speed, volume, pitch, and optional word timestamps. \\
trim-audio & Trim an audio file to a specified start and end time. \\
trim-video & Trim a video to a specified start and end time. \\
video-gen & Generate or edit videos with video-generation APIs, prompt rewriting, reference inputs, preprocessing, and iterative logging. \\
video-shot-split & Count and split video shots into separate clip files. \\
vision-understanding & Analyze image or video content and answer visual understanding questions. \\
\bottomrule

\end{tabular}

\end{table}

\FloatBarrier

\clearpage
\section{Benchmark Execution Costs}
\label[appendix]{app:execution_costs}

We report the cost in USD for three representative harness/model combinations used in our main experiments. For Claude Code + Opus 4.7 and Codex + GPT-5.5, we use the coding plan price, where the LLM backend cost is a fixed 200 USD subscription cost for running the full 285-case benchmark, which is equivalent to 0.7018 USD per case.

\begin{table}[htbp]
\caption{Average cost per case in USD for three representative harness/model combinations.}
\label{tab:execution_costs}
\centering
\small
\setlength{\tabcolsep}{5pt}
\begin{tabular}{lrrrr}
\toprule
\textbf{Harness / Model} & \textbf{Image} & \textbf{Video} & \textbf{Model} & \textbf{Total} \\
\midrule
Claude Code + Opus 4.7 & 0.0194 & 9.3016 & 0.7018 & 10.0228 \\
Claude Code + DeepSeek & 0.0633 & 9.2084 & 0.1853 & 9.4570 \\
Codex + GPT-5.5 & 0.0244 & 9.7484 & 0.7018 & 10.4746 \\
\bottomrule
\end{tabular}
\end{table}

Running one case costs approximately 9.5--10.5 USD. This cost is dominated by video-generation API calls.

\end{document}